%% file: main.tex
\documentclass[journal,comsoc]{IEEEtran}

\usepackage[T1]{fontenc}
\usepackage{cite}
\usepackage{booktabs}
\usepackage{multirow}
\usepackage{amsmath,amssymb,amsfonts,amsthm}
\usepackage{algorithm}
\usepackage{algorithmic}

\usepackage{textcomp}
\usepackage{url}
\usepackage{bbm}
\usepackage{enumerate}
\usepackage[table,xcdraw]{xcolor}
\usepackage{graphicx}
\usepackage{graphics}
\usepackage{subfigure}
\usepackage{pifont}
\usepackage{lipsum}

\usepackage{caption}
\usepackage{bbding}

\ifCLASSINFOpdf

\else
 
\fi

\usepackage{amsmath}

\usepackage[cmintegrals]{newtxmath}

\ifodd 1
 
\newcommand{\com}[1]{\textbf{\color{red} (COMMENT: #1)}} 
\newcommand{\comg}[1]{\textbf{\color{green} (COMMENT: #1)}}
\newcommand{\response}[1]{\textbf{\color{magenta} (RESPONSE: #1)}} 
\else

\newcommand{\com}[1]{}
\newcommand{\comg}[1]{}
\newcommand{\response}[1]{}
\fi

\begin{document}

\title{\huge Edge Robotics: Edge-Computing-Accelerated \\Multi-Robot Simultaneous Localization and Mapping}

\author{\IEEEauthorblockN{Peng Huang\textsuperscript{$\dagger$}, Liekang Zeng\textsuperscript{$\dagger$}, Xu Chen, Ke Luo, Zhi Zhou, Shuai Yu} \\

\thanks{$\dagger$ The two authors contributed equally to this work.}
\thanks{
The authors are with the School of Computer Science and Engineering, Sun Yat-sen University, Guangzhou,
Guangdong, 510006 China, e-mail: \{huangp57, zenglk3, luok7\}@mail2.sysu.edu.cn, \{chenxu35, zhouzhi9, yushuai\}@mail.sysu.edu.cn. 
The corresponding author is Xu Chen.}
}

\maketitle

\begin{abstract}
With the wide penetration of smart robots in multifarious fields, Simultaneous Localization and Mapping (SLAM) technique in robotics has attracted growing attention in the community. 
Yet collaborating SLAM over multiple robots still remains challenging due to performance contradiction between the intensive graphics computation of SLAM and the limited computing capability of robots.
While traditional solutions resort to the powerful cloud servers acting as an external computation provider, we show by real-world measurements that the significant communication overhead in data offloading prevents its practicability to real deployment.
To tackle these challenges, this paper promotes the emerging edge computing paradigm into multi-robot SLAM and proposes RecSLAM, a multi-robot laser SLAM system that focuses on accelerating map construction process under the robot-edge-cloud architecture.
In contrast to conventional multi-robot SLAM that generates graphic maps on robots and completely merges them on the cloud, RecSLAM develops a hierarchical map fusion technique that directs robots' raw data to edge servers for real-time fusion and then sends to the cloud for global merging.
To optimize the overall pipeline, an efficient multi-robot SLAM collaborative processing framework is introduced to adaptively optimize robot-to-edge offloading tailored to heterogeneous edge resource conditions, meanwhile ensuring the workload balancing among the edge servers.
Extensive evaluations show RecSLAM can achieve up to $39.31$\% processing latency reduction over the state-of-the-art.
Besides, a proof-of-concept prototype is developed and deployed in real scenes to demonstrate its effectiveness.

\end{abstract}


\IEEEpeerreviewmaketitle

\section{Introduction}
\label{sec:introduction}

\IEEEPARstart{A}{S} a key enabling technology in robotics, Simultaneous Localization and Mapping (SLAM) is a computationally-intensive approach that targets at simultaneously constructing a graphic map and tracking the agent's location in an unknown environment \cite{cadena2016past,huang2019survey,huang2016critique}.
It has been broadly used across a number of applications, opening a wide door to smart robots for understanding and interacting with environments wherever indoor, aerial, or underground. For example, in autonomous robotics, SLAM has been employed to process the surrounding information from laser scans to guide the self-contained navigation \cite{sun2020scalability, buyval2018realtime, fan2018baidu,badue2021self}. In disaster relief, snakelike robots apply SLAM to explore fragile buildings and carry out a rescue in human-unreachable places \cite{ruckert2021snake, xiao2018review, wang2020three}.

Generally, SLAM can be categorized into two classes according to the used sensors, namely visual SLAM\cite{taketomi2017visual} and laser SLAM\cite{cole2006using}. Visual SLAM utilizes the images captured by cameras to reckon geometric graphics.
It strives to exploit the 2D visual input to construct 3D semantic models.
Laser SLAM forks another technical path that extracts data from laser scans to shape a grid occupancy map. 
Particularly, laser data differs from visual images by containing depth-wise information and high-precision positions of objects.
Benefited by such advantage in sources, laser SLAM is preferred and employed for a much wider spectrum of robotic applications for achieving much higher physical accuracy \cite{wang2020urban}.

\begin{figure}[t]
    \centering
    \subfigure[Example apartment scene with ten robots.]{
        \begin{minipage}[t]{0.4\linewidth}
        \centering
        \includegraphics[height=3.5cm]{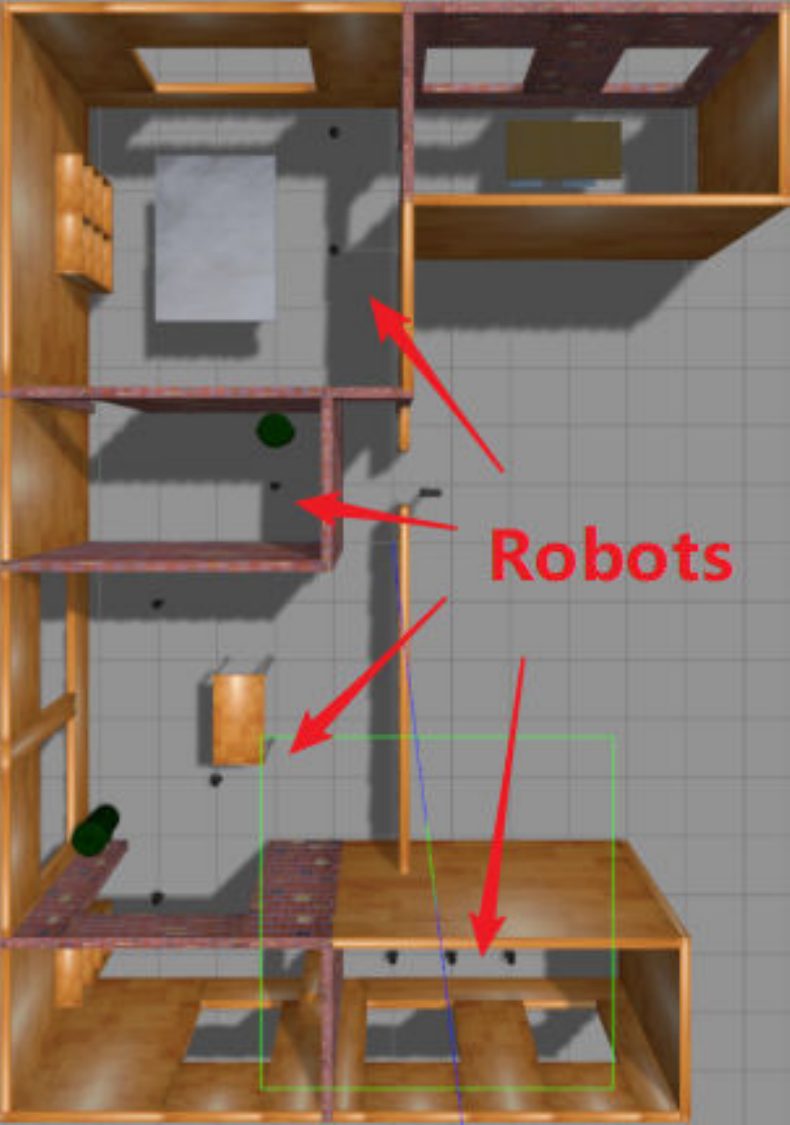}
        \end{minipage}
    }
    \quad
    \subfigure[The corresponding map generated by RecSLAM.]{
        \begin{minipage}[t]{0.4\linewidth}
        \centering
        \includegraphics[height=3.5cm]{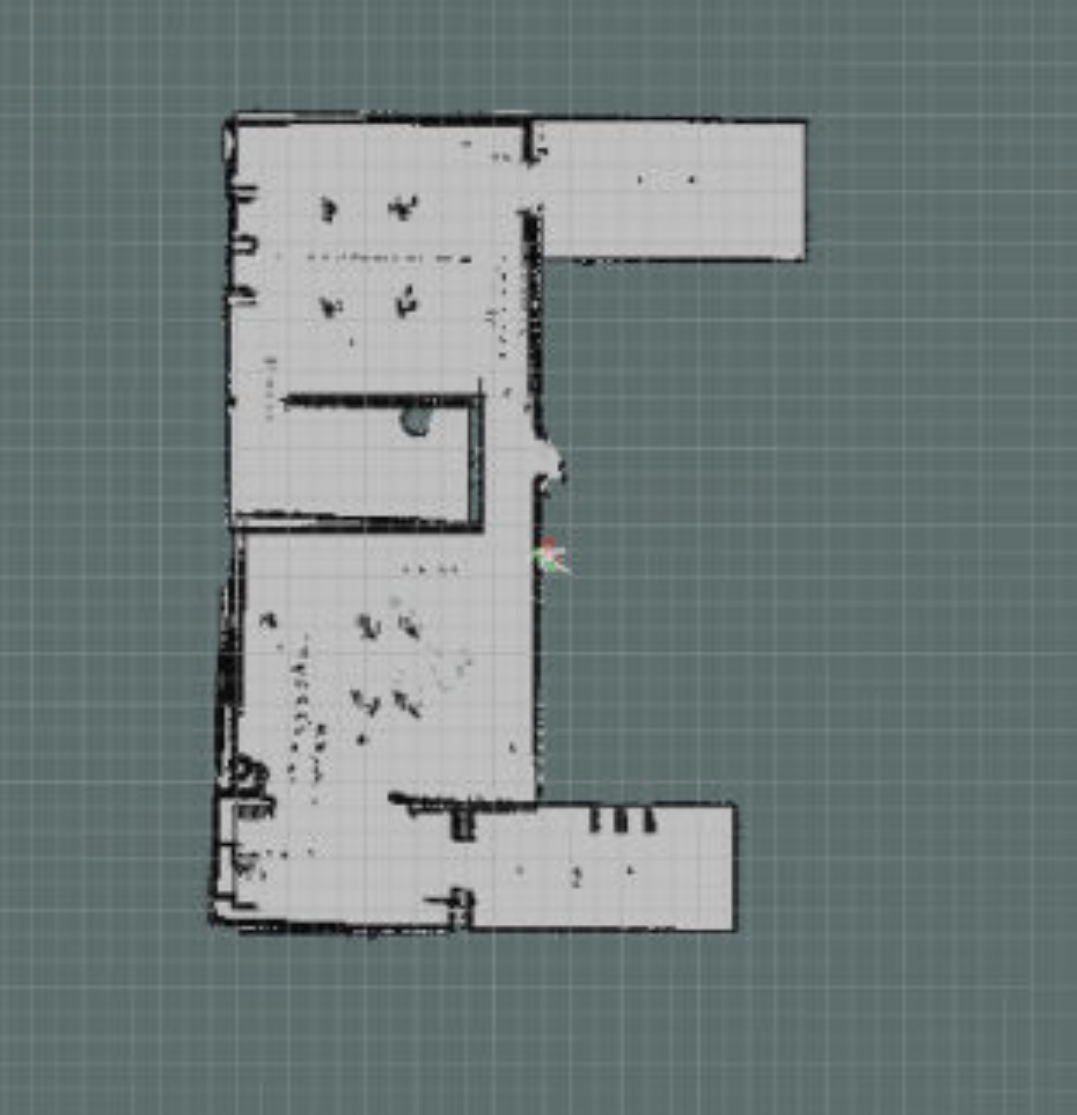}
        \end{minipage}
    }
    \caption{An example multi-robot SLAM application at the edge, where the robots are navigated to walk in the apartment (a) and perform SLAM to construct the corresponding geometric indoor map (b).}
    \label{figure:simulation}
\end{figure}

The widespread use of smart robots catalyzes SLAM to extend its deployment over multiple robots\cite{saeedi2016multiple, dube2017online, rizk2019cooperative}.
Specifically, a group of robots is deployed in a swarm manner, and they
are committed to cooperating with each other in an unknown environment in order to construct the geometric mapping and determine their own locations through SLAM.
An example scenario is to build the graphic maps of an apartment as shown in Fig. \ref{figure:simulation}, where multiple robots are employed to perform SLAM simultaneously for global map construction.
Another example is in mission-critical Search and Rescue (SAR) tasks \cite{dube2017online,saeedi2016multiple}, where multiple robots are driven collaboratively to share
map data with each other for rapid exploration in broken buildings.
In these circumstances, traditional wisdom \cite{zhang2018cloud, sarker2019offloading, yun2017towards} to reconcile the distributed data sources calls for robot-cloud synergy, where robots run SLAM individually and upload their local map mutually to a centralized cloud server for global merging.
Yet the two-tier architecture suffers from dual drawbacks.
On the one hand, processing SLAM workloads locally on commodity robots typically demand plenty of latency and can even fail to complete under resource constraints, \textit{e.g.} limited memory capacity.
In our measurements on an ordinary robot (as will be shown in Sec. \ref{sec:implication}), running SLAM can force CPU workload at a high level consistently (always >83.60\%), leading to excessive latency in on-device execution.
Such significant and excessive latency impedes many multi-robot applications (e.g., disaster relief and high-resolution autonomous navigation) that generally demand real-time information fusion. On the other hand, transferring robots' local maps of a large volume to the remote cloud through the unreliable and delay-significant wide-area Internet connection not only incurs the communication bottleneck but raises users' concerns on security and privacy.
Quantitatively, as we will show later, the data uploading time overhead can dominate the whole processing in common 4G, 5G and WiFi networks, exhibiting vulnerable performance.

To remedy these limitations, this paper intends to leverage the emerging edge computing \cite{shi2016edge, zhou2019edge, deng2020edge} paradigm to perform multi-robot laser SLAM in low latency.
Instead of relying on geographically distant data centers, edge computing concentrates on utilizing vicinal computing resources (e.g., 5G/WiFi edge severs) in physical proximity to end devices, and therefore remarkably shortens data communication distance, lowers offloading transmission delay, and allows the advanced quality of services \cite{zeng2020coedge, ouyang2019cost}.
Nevertheless, enabling multi-robot SLAM with edge computing architecture is non-trivial with three-fold challenges.
First, the spatially distributed nature of multi-robot deployment requires a unified flow for data collection, organization, and computation due to the versatile sources as input.
For instance, to collect data from two physically isolated robots, the edge server needs to identify their source location, timestamp, and sensory format (e.g., raw data from whether laser scan or IMU sensors).
Second, distributing the map fusion, which is traditionally centralized processed, to multiple edge servers desires a retrofit on handling SLAM maps, which is intractable under resource dynamics.
Third, simultaneously orchestrating the computation and communication among multiple robots and multiple edge servers presents inherent complexity in formulation and optimization on SLAM execution and data transmission.
Furthermore, the network dynamics among robots and edge servers further complicate the problem.

To tackle these challenges, this paper proposes RecSLAM, a system built upon the \textbf{R}obot-\textbf{e}dge-\textbf{c}loud architecture to enable multi-robot laser \textbf{SLAM} in low-latency services. 
RecSLAM's design is motivated by the observations on real-world measurements that 1) migrating SLAM workloads from robots to edge servers can effectively augment the robots' processing capability, and 2) preparative merging a subset of local maps at the edge can shrink the sizes of data to be uploaded to the cloud and therefore further reduces communication costs.
By exploiting them, RecSLAM considers a hierarchical robot-edge-cloud pipeline, where each robot individually collects data and selectively transfers them to one of the edge servers.
The edge servers commit to performing SLAM on their received robot data, merging the corresponding local maps, and uploading the preparative fusion results to a dedicated cloud server for global map fusion.
An adaptive coordinator (i.e., task scheduler) is further developed to optimize the data flow between robots and edge servers, aiming at minimizing the end-to-end latency.
We implement RecSLAM on both simulation platforms and real robots.
Simulation results show that our system can outperform the state-of-the-art solutions by up to 39.31\% latency reduction. 
The proof-of-concept prototype in an indoor scene corroborates its feasibility and efficiency, demonstrating the promising advantage of edge-accelerated SLAM.
In summary, we make the following key contributions:

\begin{itemize}
\item We conduct a fine-grained investigation on the processing costs of existing SLAM processing solutions.
The robotics-based measurements reveal that the cloud offloading mechanism suffers from the considerable transmission latency in sensory data uploading, while the local execution falls short at efficiency due to the limited onboard computing resource. 
\item We propose RecSLAM, a collaborative SLAM system based on hierarchical robot-edge-cloud architecture to enable real-time SLAM serving.
RecSLAM decouples the conventional SLAM pipeline to distribute them to multiple edge servers, where each robot can selectively offload their raw frames to one of the edge servers and the edge server can fuse multiple frames ahead of global merging.
By extending the centralized computation to distributed and parallel processing, RecSLAM significantly improves the utilization of edge resources.
\item We develop a novel graph based framework to optimize the collaborative SLAM processing between robots and edge servers.
Specifically, we separate the workflow into two stages, namely robot data grouping and edge offloading.
For robot data grouping, we abstract an undirected graph to describe the distributed robot data processing issue with workload-balancing among the edge servers, and apply an efficient balanced graph partitioning algorithm to make a balanced robot data grouping for offloaded processing at the edge.
For edge offloading, we devise an efficient resource-aware collaborative processing strategy to adaptively offload the grouping data from the robots to the proper edge servers, in order to minimize the total processing latency.
\item We implement and evaluate RecSLAM in both simulation and realistic deployment.
The simulation on the Gazebo\cite{koenig2004design} platform demonstrates the effectiveness of RecSLAM and its collaborative processing algorithms, showing up to 39.31\% latency speedup upon cloud offloading approach.
The realistic prototype on three robots in an indoor experimental scene verifies its feasibility and validity in rendering efficient SLAM services for edge applications.
\end{itemize}

The rest of this paper is organized as follows.
Section \ref{sec:motivation} characterizes the existing multi-robot SLAM solution.
Section \ref{sec:design}, \ref{sec:optimization} and \ref{sec:implementation} present the design, optimization and implementation of the proposed system.
Section \ref{sec:evaluation} evaluates in terms of both simulation and prototype.
Section \ref{sec:related_work} reviews related works and Section \ref{sec:conclusion} concludes.

\section{Background and Motivation}
\label{sec:motivation}

This section dives into multi-robot SLAM processing by characterizing the state-of-the-art cloud-based solution.
We first briefly introduce its workflow, and next break down the component-wise performance, which reveals the opportunities and challenges of edge computing.

\subsection{Multi-Robot SLAM}
\label{sec:slam}

\begin{figure}[t]
\centerline{\includegraphics[width=0.95\linewidth]{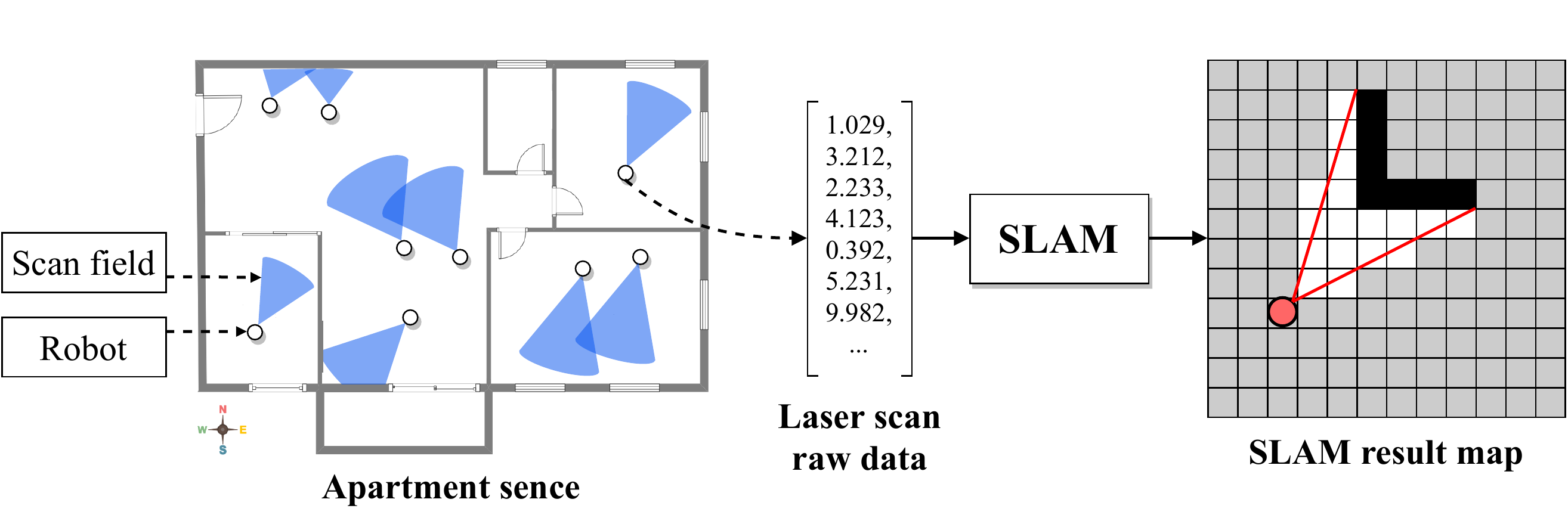}}
\caption{The illustration of a typical SLAM computation. The example scenario covers several robots that run simultaneously to sketch an indoor map for the apartment, where each robot is represented by a dot and its scan field is abstracted as a sector. A robot walks intentionally to acquire the environment's laser scan data, in the form of a floating-point vector, and passes it to the local SLAM computation module to generate a geometric grid map.}
\label{fig:SLAM_input_output}
\end{figure}

Simultaneous Localization And Mapping (SLAM) is typically run at a robot with various sensors equipped, aiming at simultaneously localizing the robot itself and maintaining a graphic map of the environment \cite{schreck2021slam}.
Fig. \ref{fig:SLAM_input_output} shows a typical scenarios of SLAM computation.
The input to SLAM is the sensory data from robots in a form of floating-point vectors, which describe the scanning angle, laser distance, etc.
The output is an occupancy grid map, where the black grids are obstacles, the whites represent available safe space and the grays mean the places that have not been detected or uncertain.

\begin{figure}[t]
\centerline{\includegraphics[width=0.98\linewidth]{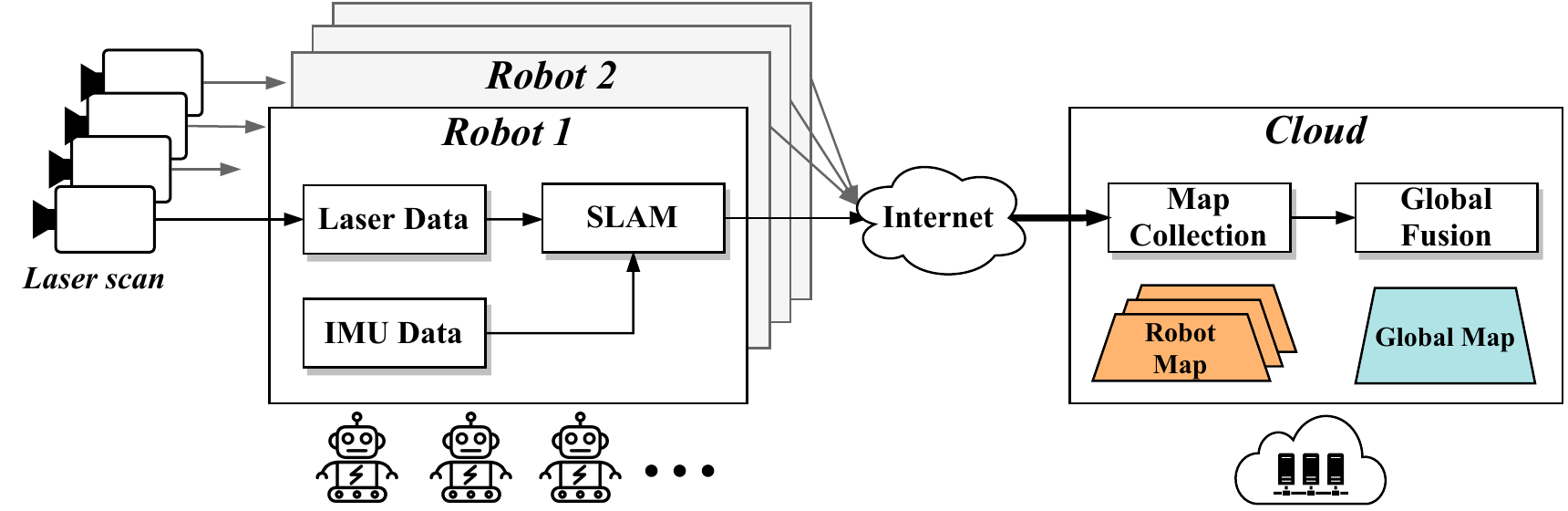}}
\caption{The pipeline of the traditional cloud-based multi-robot SLAM solution, where the robots performs SLAM individually and upload their data via Internet to a centralized cloud server for global map fusion.}
\label{fig:cloud_processing}
\end{figure}

Multi-robot SLAM is an extended use case of single-robot SLAM, which manages a cluster of robots to perform SLAM for constructing a virtual graphic sense corresponding to reality \cite{gautam2012review}.
Its core procedure beyond single-robot SLAM is the map fusion procedure that merges multiple grid maps to yield a global map of the targeted scene.
Given that grid maps are collected from multiple distributed robots, state-of-the-art solutions \cite{zhang2018cloud, sarker2019offloading, yun2017towards} resort to the centralized cloud server to fulfill map fusion, as illustrated in Fig. \ref{fig:cloud_processing}.
Particularly, it works in two stages.
First, each robot collects sensory data and individually performs SLAM locally.
Next, they upload their local SLAM result, i.e., grid maps, to the cloud, where all these maps are merged to acquire a global map.
Specifically, two local maps that share an overlapping area will be fused, while those independent ones will be simply appended to constitute the global map.
This global map can be used for downstream tasks such as navigation and furniture design.

Multi-robot SLAM has been widely adopted in a broad range of scenarios.
For example, some indoor mapping services employ multiple robots to scan the house and perform multi-robot SLAM to construct a 3D graphic model that shapes the indoor environment \cite{dube2017online}.
In some geographical mapping applications, many robots are run to scan landforms and build topographic maps \cite{zhang2018cloud}.

\subsection{Performance Implications of State-of-The-Art}
\label{sec:implication}

\begin{figure}[t]
\centerline{\includegraphics[width=0.65\linewidth]{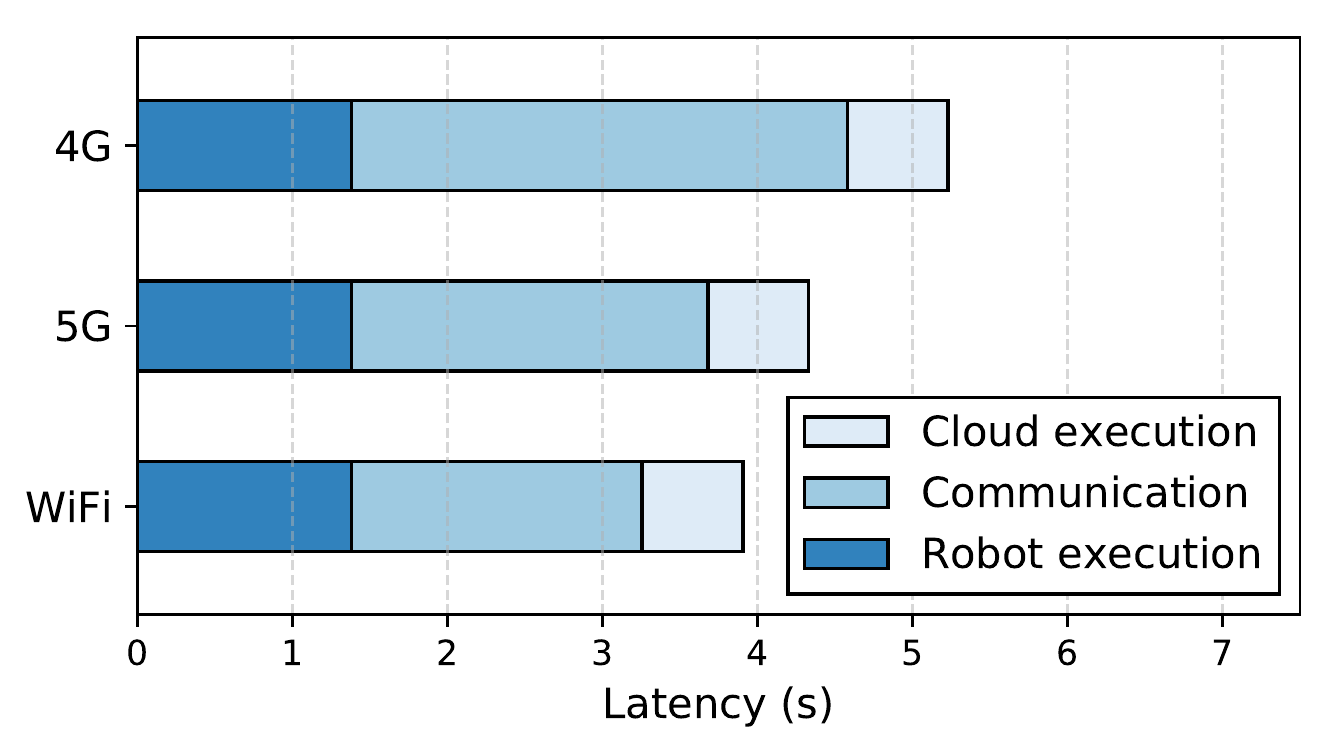}}
\caption{The latency breakdown of a cloud-based multi-robot SLAM solution. The communication latency contributes a major portion out of the total and is sensitive the networking condition. }
\label{fig:cloud_cost}
\end{figure}

\begin{figure}[t]
\centerline{\includegraphics[width=0.7\linewidth]{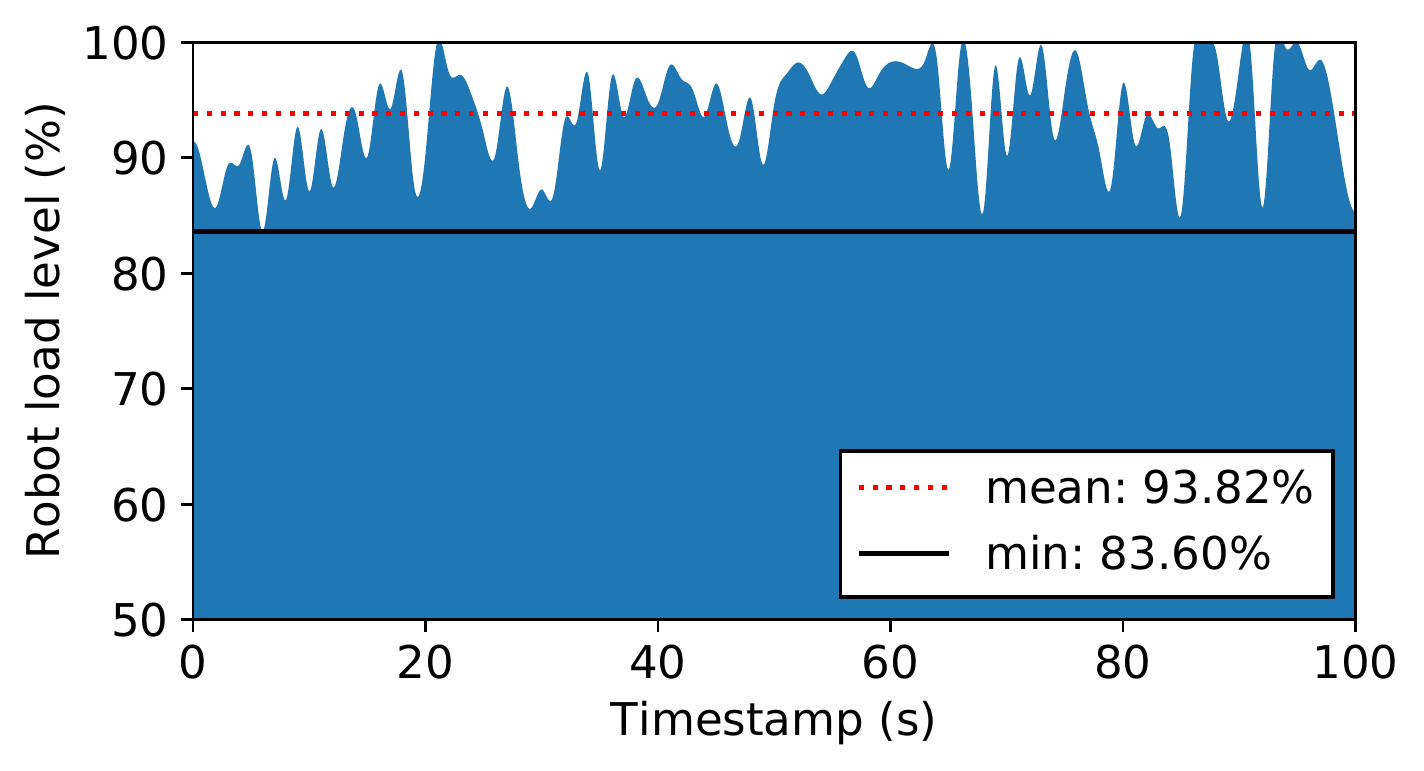}}
\caption{The monitored background load level during SLAM runtime on the robot. The load curve consistently lies at a high level with a minimum of 83.60\% and an average of 93.82\%, indicating the robot is overloaded.}
\label{fig:motivation_cpu_workload}
\end{figure}

State-of-the-art cloud-based solutions highly rely on the Internet to gather local frames from distributed robots, which makes it sensitive to the vulnerable network.
To make a clearer understanding of how the network conditions impact, we explicitly examine the costs of a typical cloud serving process.
Specifically, we deploy a multi-robot SLAM prototype using three Turtlebot3 and a cloud server, aiming at measuring the duration from laser scan input at robots to a global map obtained on the cloud.
The Turtlebot is equipped with Raspberry Pi as the processing core, which has 1.4GHz ARM Cortex-A53 CPUs, 1GB LPDDR2 without GPU. The cloud server is with Intel Xeon CPU E5-2678, deployed at the available region that robots locate.
The robots contact the cloud via three channels: 3G, 4G, and WiFi, all are under commercial operation networks.

\begin{figure*}[]
\centering 
\includegraphics[width=0.85\textwidth]{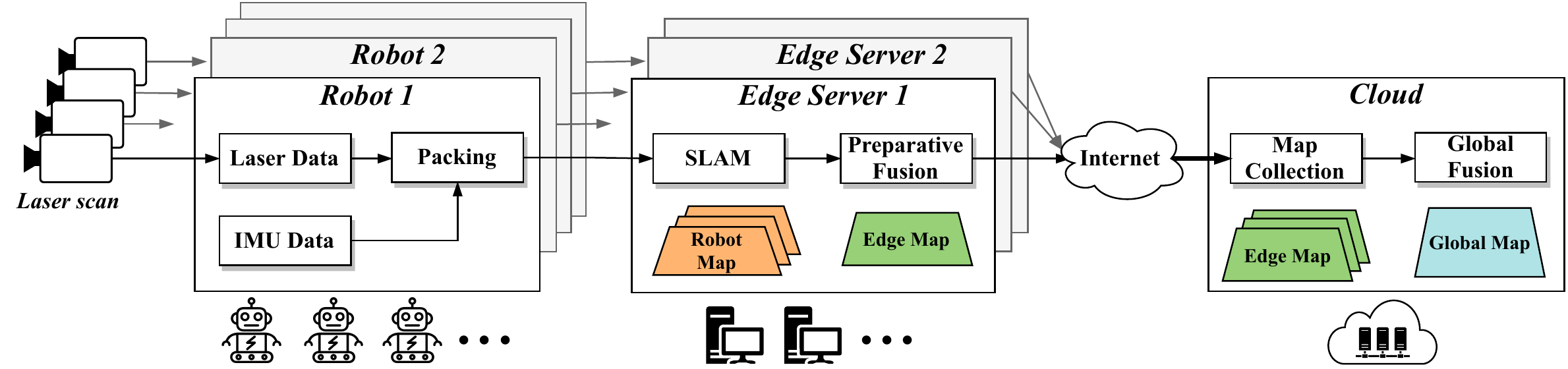}
\caption{RecSLAM architecture overview. Each robot individually collects its sensory data and directly packs and transfers to a dedicated edge server.
Each edge server runs SLAM using the received data and performs a preparative fusion to aggregate an edge map with its owned robot maps.
All the edge maps will be sent to a centralized cloud for global map fusion.}
\label{fig:overview}
\end{figure*}

Fig. \ref{fig:cloud_cost} visualizes the measurements, where the latency breaks down in robot computing, frames uploading, and cloud merging.
We can witness that the total latency is highly sensitive to the channel switching with an increase from about 3.83s (WiFi) to 5.36s (5G).
Looking closer, we observe that such cost fluctuation comes from the communication side, which dramatically varies under different network conditions.
Concretely, the uploading latency takes 3.13ms, 2.35ms, 1.88ms for 4G, 5G, and WiFi, respectively.

Another expensive workload revealed is the local computing cost.
To compute SLAM at the robot consumes around 1.51s, occupying a percentage of 49.74\% in total latency under WiFi channel.
To inspect the cause behind that, we log the background load level on robots and plot the trace as in Fig. \ref{fig:motivation_cpu_workload}.
As shown in the figure, the trajectory of the load always goes on top of 80\%.
It records an average of 93.82\% and a minimum of 83.60\%, sometimes even reaching 100.00\%, indicating that the robot is overloaded and can not sufficiently support the local SLAM computation.

In summary, the existing cloud-based solution suffers from dual bottlenecks.
One is the inherent communication bottleneck from the remote Internet transmission that stresses the local frame uploading process between robots and the cloud.
The other is the resource-hungry SLAM computation tasks, conflicting with the limited computing capability of the robots.
The integration of these two knobs makes the cloud serving fall short in rendering low latency multi-robot SLAM, desiring a holistic retrofit on the execution pipeline.

\subsection{Opportunities and Challenges with Edge Computing}
To tackle the knobs of cloud-based approaches in both data offloading and robot computing, we promote edge computing to enable multi-robot SLAM processing.
Edge computing sinks computing power to the physical proximity to the robots and thereby provides a potential assist to greatly reduce the computation stress for robots while significantly lowering the transmission latency compared to cloud offloading paradigms \cite{zhou2019edge}.

With edge computing, there are opportunities to redesign the multi-robot map fusion pipeline to accomplish both efficient SLAM computation and low latency transmission.
Specifically, we can utilize the distributed edge servers to take over the workload from both robots and the cloud, acting as agents to resolve the processing bottlenecks smoothly.

Nevertheless, different from the cloud data center that provides unified, powerful, and closed resource access, edge servers exhibit a loosely coupled and uneven nature: they are usually distributed, heterogeneous, and dynamic.
To orchestrate the serving between robots, edge servers, and cloud, the communication protocol should be carefully designed considering data transmissions in robot-to-edge, edge-to-edge, and edge-to-cloud.
Besides, the workload balance among the multiple edge servers, as another impacted aspect related to the overall performance, should also be taken into account given the network dynamics.

\section{RecSLAM System Design}
\label{sec:design}

In this section, we introduce the design details of RecSLAM, a multi-robot laser SLAM system that leverages hierarchical robot-edge-cloud architecture to accelerate multi-robot map construction. 
We present the design of RecSLAM by explaining its modular details following the data flow.

\subsection{Overview}
The environment we envisaged to deploy RecSLAM spans multiple robots with laser scans and multiple edge servers in proximity.
The robots can walk randomly or along with a preset route, and therefore their sensory data are time-varying and may be overlapped in some moments.
The edge servers can accept data from robots simultaneously and promise to be available during an epoch of execution.

RecSLAM works upon robot-edge-cloud architecture and designs specific modules in each tier.
Fig. \ref{fig:overview} shows a high-level view of RecSLAM.
In the beginning, each robot captures environmental perceptions via its installed sensors and acquires the raw sensory data.
In RecSLAM, we mainly consider two kinds of typical data: laser data from the laser scans and acceleration data from the Inertial Measurement Unit (IMU).
Instead of running SLAM locally in a traditional mechanism, we immediately pack the sensory data and offload it to a certain edge server.
When it accepts a data package, the edge server launches the SLAM execution and obtains a map corresponding to it.
For clarity, we identify this map as \textit{local map} or \textit{robot map}, indicating a direct mapping to a robot's raw data.
For an edge server that serves multiple robots, a sequence of the local maps are generated and will be merged in situ to an \textit{edge map}.
This merging process is referred to as \textit{preparative fusion}, as it is a preparatory action towards \textit{global fusion} in the cloud.
All the edge maps from distributed edge servers are finally merged in a centralized cloud to calculate the global map.

\subsection{Robots: Data Collection and Packing}
As discussed in Section \ref{sec:implication}, one of the performance bottlenecks is the excessive computing latency on robots, which comes from the contradiction between resource-hungry SLAM computation and resource-constrained on board processors.
To alleviate this, we propose to reserve only the lightweight data collection procedure on robots.
Specifically, in our implementation, the sensory data to be collected are laser data and IMU data.
The former is a structure named \textsf{sensor\_msgs::LaserScan}, including timestamp, range data, angular distance and so on, while the latter contains information for coordinates' transformation.
Though our current deployment only takes these two types of sensory data into account, it is convenient to accommodate additional formats in RecSLAM.

To offload the SLAM computation, we pack these raw data and transfer them to edge servers through Robot Operating System (ROS).
Particularly, each robot establishes a connection and offloads its data to exactly one edge server, while each edge server can serve multiple robots at the same time.
Here we assume that these connections are relatively well and stable, which can be relevant \cite{ben2020edge,dey2016robotic} provided that they are within the same LAN.
Therefore, the mobility of robots will not influence the data transmission and the consequent procedures.
We note that in realistic deployment the robot-edge communication may be fluctuated and even fail, which raises robustness issues in multi-robot SLAM system. 
To address that, we can leverage multi-access mobile edge computing \cite{wu2018noma,wu2019delay,wu2018optimal} to enable partial offloading, where a robot can transfer its data to multiple edge servers to ensure service availability and improve system performance.

\subsection{Edge Servers: Preparative Fusion for Edge Maps}
The robots can access a set of edge servers in proximity via 5G/WiFi connections, and the edge servers, take charge of the SLAM computation and assisting the map fusion.
Each edge server accepts raw data from multiple robots nearby and sends its result to a cloud.

The SLAM computation is inherently tightly coupled.
If it is forced to split, the explosive complexity would lead to additional overhead in handling the graphic algorithms.
Therefore, we treat the SLAM computation as a whole by encapsulating it as a function module for processing the raw data from a robot. In this case, the input to SLAM is collected from robots, and the output will be passed to map fusion.
In our prototype, we develop the SLAM module with GMapping \cite{grisetti2007improved}, a popular lightweight open-source SLAM implementation.
For further acceleration, we manage a pool of execution instances to enable parallel and asynchronous serving, i.e., launch a separated process to compute SLAM whenever a robot's raw data arrive.

The results of these processing threads are called robot maps, which will be merged via a fusion module.
We refer to this module as preparative fusion, in contrast to the global fusion on the cloud.
Particularly, the preparative fusion targets at the robot maps on the edge server and perform fusion by first checking the overlapping degree of robot maps and next merging those overlapping ones.
We design to separate such a preparatory process from the global fusion for two reasons.
On the one hand, by sinking the fusion workload to the edge servers, we can shrink the computing latency by avoiding excessive delay-significant transmissions to the remote cloud.
On the other hand, fusing the robot maps on an edge server can effectively deduplicate the redundancy across overlapping robot maps and thus reduce the data size of edge maps, for which the communication overhead between edge servers and cloud is saved.

\subsection{Cloud: Aggregated Fusion for Global Map}
\label{sec:design_cloud}

The cloud is responsible to collect all edge maps from edge servers.
We designate the cloud as the destination of final merging because a large number of downstream applications of SLAM are deployed on the cloud.
However, if some service needs to deploy a SLAM-based task on end devices, we can expediently assign an edge server to finish global fusion and push the result to the specific device.

In the robot-edge-cloud architecture of RecSLAM, each edge server runs its SLAM and preparative fusion individually as long as it receives the robot data, which implies a parallel and distributed processing paradigm among the edge servers.
To optimize the performance in such workflow, we need to carefully decide on a robot data processing assignment strategy with a target of balanced workload distribution among the edge servers to avoid the straggler effect.
Nevertheless, this problem is non-trivial in its two-fold challenges.
First, the overlapping degrees (i.e., information redundancy) across the robot maps (wrt. the robot's sensory data) vary greatly among the robots given the mobility and randomness of the robots' walk, while they can largely impact the computing latency of SLAM on edge servers and the data size of the fused edge maps.
Both these two metrics are key performance contributors towards total execution time.
Second, the network conditions between the robots and cloud can be heterogeneous.
This requires the robot data processing offloaded from the robots to the edge servers to be aware of the diverse network connections.
To overcome these challenges, we design an efficient optimization framework for robot-edge collaborative processing in the next section.

\begin{figure*}[t]
    \centering
    \subfigure[The computing time of the map fusion procedure in multi-robot SLAM with varying number of input maps.]{
        \begin{minipage}[t]{0.25\textwidth}
        \centering
        \includegraphics[height=4.5cm]{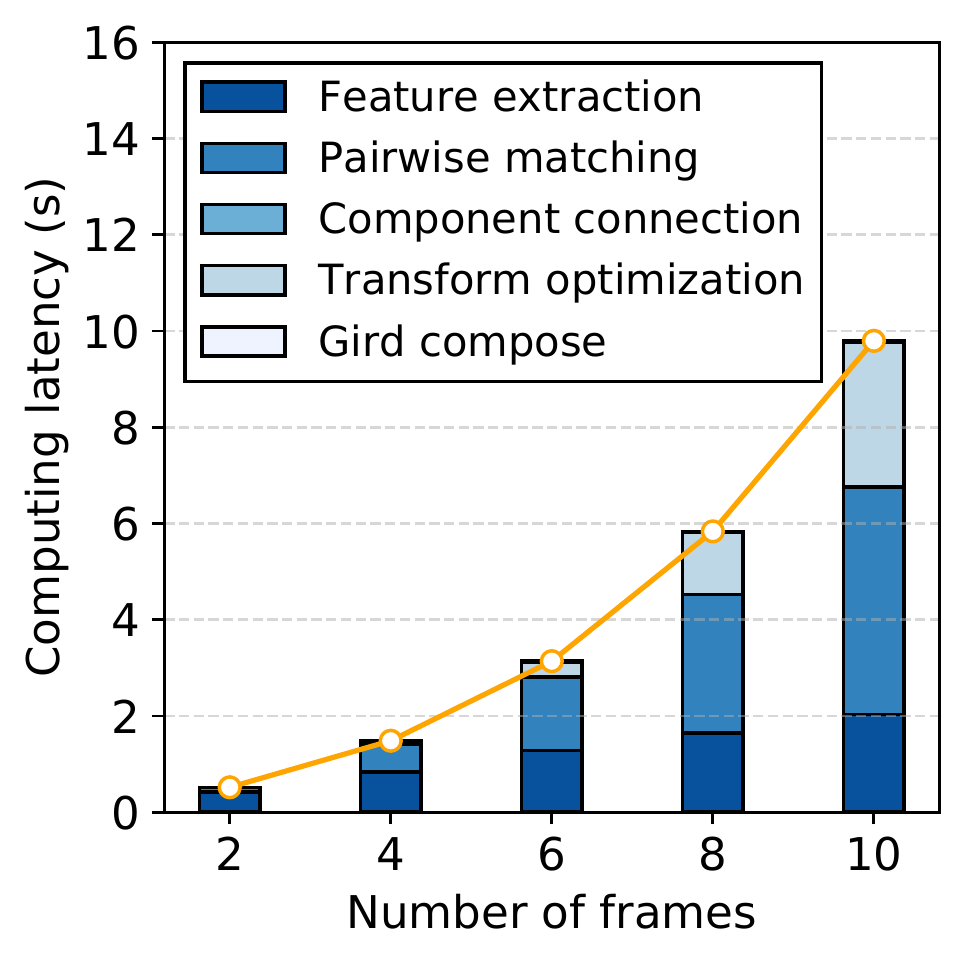}
        \label{fig:latency_break_down}
        \end{minipage}
    }
    \quad
    \subfigure[The data size of two input maps and the corresponding merged map with varying overlapping degrees.]{
        \begin{minipage}[t]{0.25\textwidth}
        \centering
        \includegraphics[height=4.5cm]{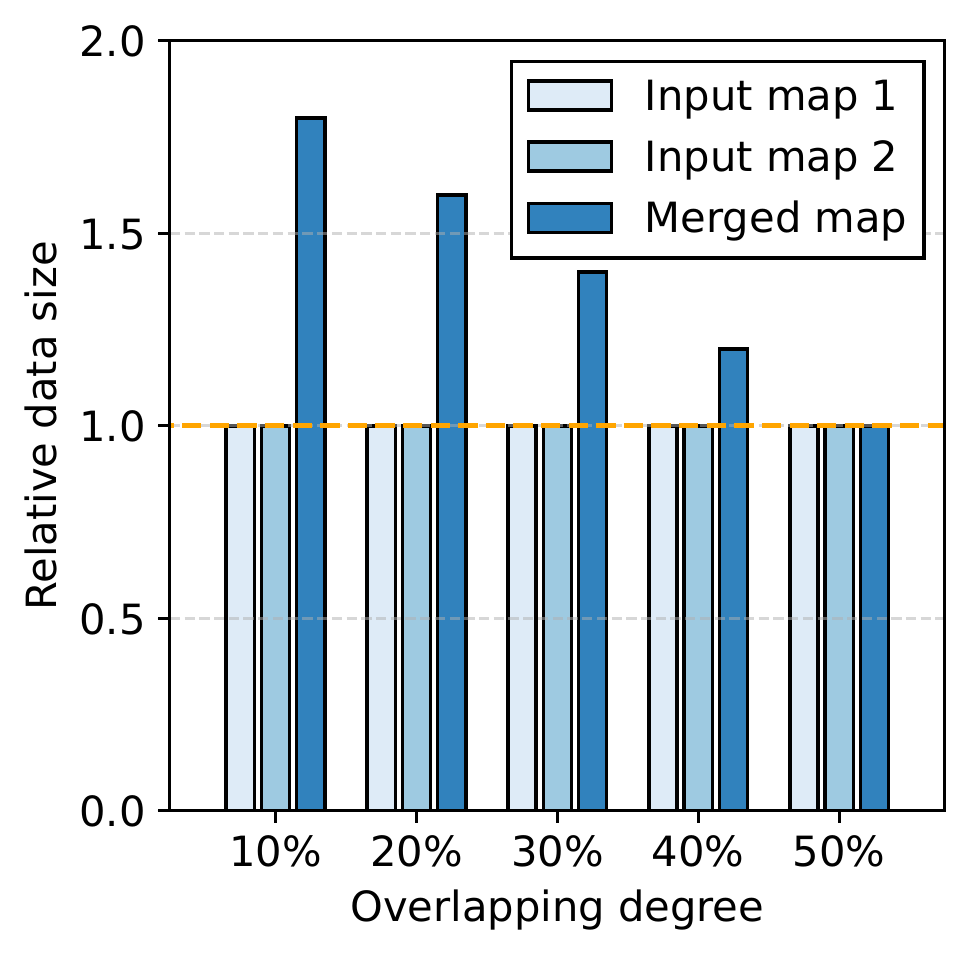}
        \label{fig:merging size}
        \end{minipage}
    }
    \quad
    \subfigure[Illustration of the map fusion procedure with two example input maps. The blue area indicates the overlapping part of the inputs.]{
        \begin{minipage}[t]{0.4\textwidth}
        \centering
        \includegraphics[height=4.5cm]{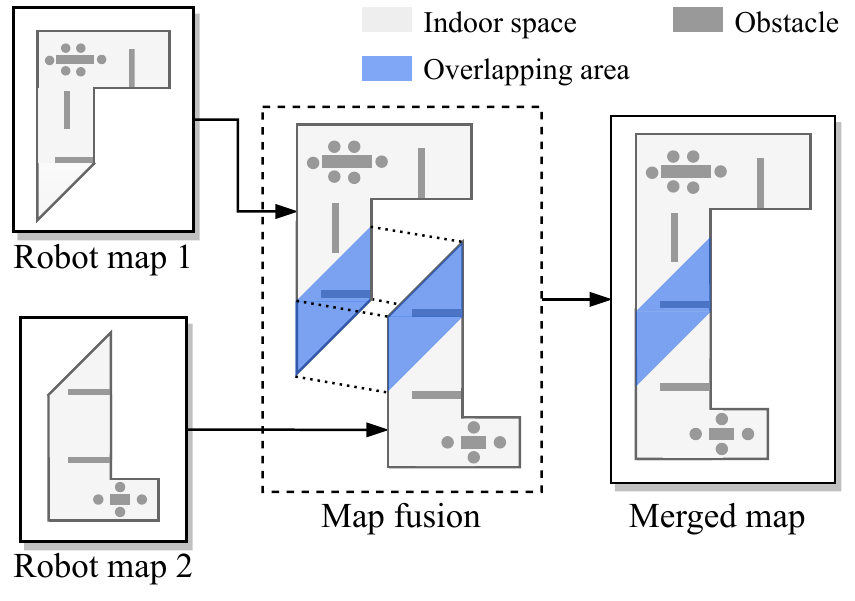}
        \label{fig:overlap_merge}
        \end{minipage}
    }
    \caption{Analysis of the map fusion procedure. 
    The subfigure (a) shows the computing time of map fusion procedure with respect to the five functions that it applies: feature extraction, pairwise matching, component connection, transform optimization, and grid composing. 
    The subfigure (b) observes that as the overlapping degree increases, the data size of the merged map decreases. Note that the two inputs are of equal data size, and the overlapping degree is defined by the percentage of the overlapping area out of the total areas.
    The subfigure (c) illustrates a map fusion procedure with two robot maps, where it first checks the overlapping area of the inputs via their feature information, and next connects these overlapping parts and constructs a merged map. The larger the overlapping area, the smaller the data size the merged map has.}
    \label{fig:map_fusion_analysis}
\end{figure*}

\section{Robot-Edge Collaboration Optimization Algorithms}
\label{sec:optimization}

This section concentrates on scheduling the data flow between robots and edge servers in order to minimize end-to-end multi-robot SLAM latency.
We first formulate the cost optimization problem in terms of the latency expired by robots, edge servers and cloud server.
Next, we address it via a two-step workflow, i.e., balanced robot data grouping and network-aware edge offloading.

\subsection{Problem Formulation}
We consider a cluster of $R$ robots, where each robot $r$ can walk freely or with the user-defined routes, and pushes its sensory data to a specific edge server $e$ among the $E$ edge servers.
To obtain a robot $r$'s robot map processed with the assist of an edge server $e$, it counts the time for data packing, transferring and frame transformation on $e$:
\begin{align}
    t_{r} = t^{\text{pack}}_{r} + t^{\text{trans}}_{r\rightarrow e} + t^{\text{frame}}_{e}. \label{eq:time_robot}
\end{align}

Moving forward the data flow, calculating the edge map on an edge server $e$ needs to wait until all its owned robot maps ready and fuse them in situ, which takes the time:
\begin{align}
    t_{e} = \max_{r\in e}t_{r} + t^{\text{fuse}}_{e}, \label{eq:time_edge}
\end{align}
where $r \in e$ means the robots whose data are assigned to the edge server $e$, $t^{\text{fuse}}_{e}$ represents the fusion latency in edge server $e$. 
From a cloud view, it strives to aggregate all the edge maps from edge servers to reckon the global map, which takes the time of receiving all edge maps and performing global fusion:
\begin{align}
    t_{c} = \max_{e}(t_{e} + t^{\text{trans}}_{e \rightarrow c}) + t^{\text{fuse}}_{c}. \label{eq:time_cloud}
\end{align}

Essentially, $t_{c}$ is the total execution time that characterizes the duration from raw data collection to global map completion, which is exactly the optimizing objectives of RecSLAM. $t^{\text{fuse}}_{c}$ represents the fusion latency in cloud. 
Therefore, we can derive the execution latency optimization problem as:
\begin{align}
    \min & \quad t_{c}, \label{eq:problem} \\
    \text{s.t.} & \quad  (\ref{eq:time_edge}), (\ref{eq:time_cloud}). \notag
\end{align}

\textbf{Discussion.}
The formulation provides an abstracted form of the problem with various details hidden.
Among them, we mainly consider two factors.
One is the computation side, where we expect the assignment from robots to edge servers is relatively even such that the effect of parallel processing is maximized.
The rationale behind can be seen in Fig. \ref{fig:latency_break_down}, where the total computing time of map fusion grows quadratically as the number of frames linearly increases.
In Fig. \ref{fig:latency_break_down}, we further break down map fusion procedure into five successive functions: \textit{feature extraction} extracts a frame' edge features, and next \textit{pairwise matching} performs feature points matching between different data frames; \textit{component connection} and \textit{transformation optimization} functions intend to solve the pose, and finally \textit{grid compose} is performed to fuse the frames.
We observe from function-wise latency that this trend primarily comes from the pairwise matching function, which performs $k(k-1)/2$ times of pair-wise checking given $k$ frames.
Therefore, if excessive robot data is assigned to a few edge servers, the imbalance workload distribution will lead to a much higher completion time of the total process.
Reflecting on Equation (\ref{eq:time_cloud}), it is for minimizing $\max_{e}(t_{e} + t^{\text{trans}}_{e \rightarrow c})$.

The other factor is the communication side that focuses on the data size of edge maps.
This metric, as we observe in measurements, is highly sensitive to the overlapping degree, i.e., the ratio of the overlapping area to the total area of input maps.
To illustrate that, Fig. \ref{fig:merging size} plots the data size of the corresponding edge map with varying overlapping degree and Fig. \ref{fig:overlap_merge} depicts how map fusion works.
In Fig. \ref{fig:merging size}, we take two robot maps of equal data size as input, and control the content of them to adjust the overlapping area.
The result reports that as the overlapping degree increases, the edge map's data size shrinks.
Particularly, a 50\% overlapping degree of two inputs indicates they are fully overlapped, and thus the obtained edge map is exactly one of the inputs and has the same data size of a single robot map.
Fig. \ref{fig:overlap_merge} explains this impact visually, where the blue area indicates the overlapping part.
The larger the overlapping area, the higher the overlapping degree, and thus the smaller data size the edge map has.
This observation drives us to assign robot maps with as higher overlapping degree as possible together, so as to yield smaller edge maps and save the transmission cost between edge servers and the cloud.

Overall, accommodating both computation and communication aspects to optimize the problem (\ref{eq:problem}) is inherently complicated.
Essentially, it is hard to solve the optimal solution fast due to its combinatorial nature.
To achieve efficient solving, we intend to decouple it into two sub-problems by separately accounting for robot data grouping and robot data offloading. On the one hand, considering the characteristics of map fusion performed by SLAM, robot data grouping for workload-balancing needs to ensure that the number of nodes in each group tends to be uniform, while the in-group overlapping degree is as large as possible, so as to avoid the straggler effect due to imbalanced processing workload assignment and reduce the total size of fused edge maps. On the other hand, bandwidth conditions of the edge servers may be different, so network-aware offloading optimization for efficient robot-to-edge data transmission and computation is also desired.

\begin{figure}[t]
    \centering
    \subfigure[An example smart factory scene with multiple robots, where some robots, e.g., robot \textit{A} and \textit{B}, have overlapping scan fields.]{
        \begin{minipage}[t]{\linewidth}
        \centering
        \includegraphics[width=0.8\linewidth]{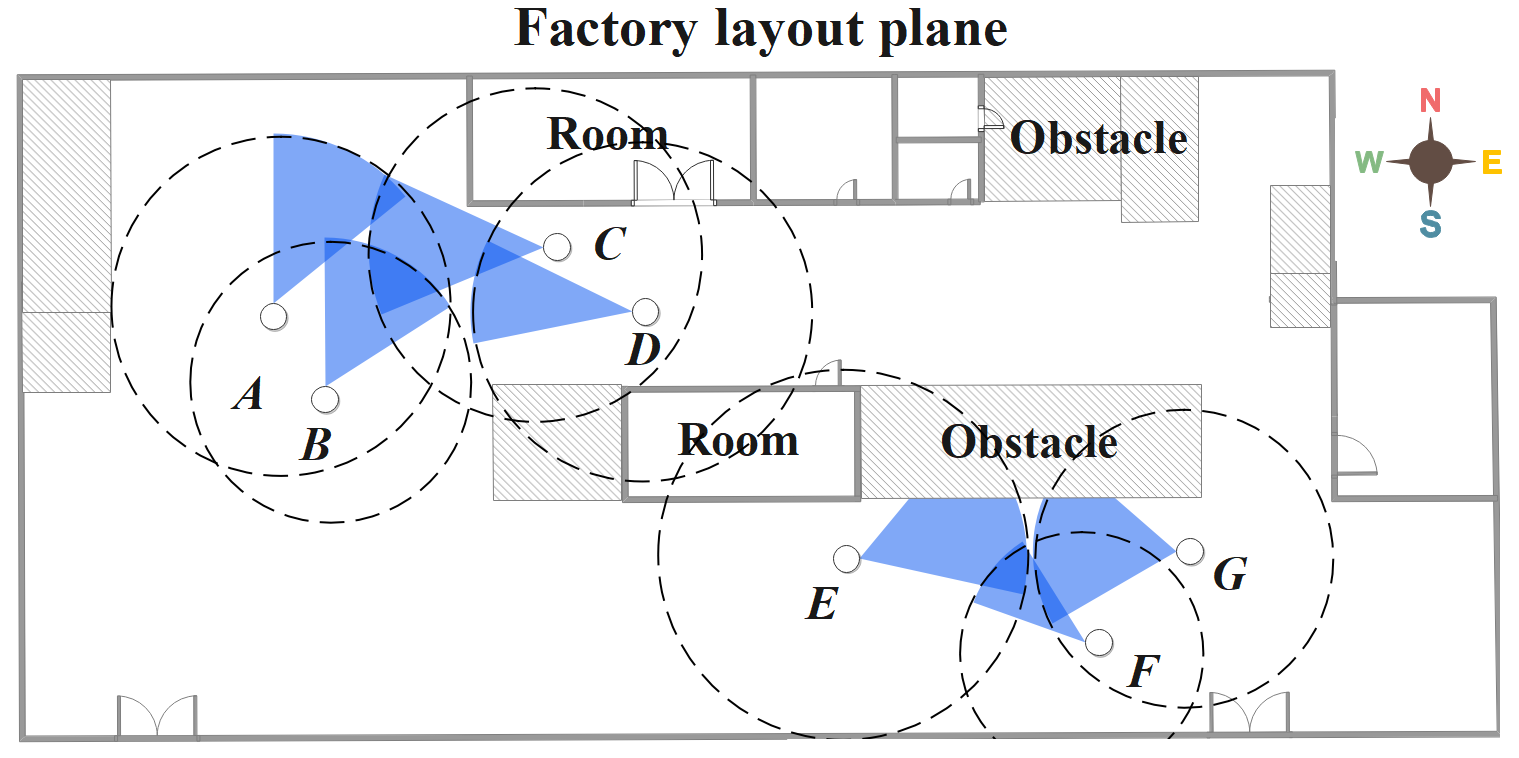}
        \end{minipage}
    }
    \subfigure[The abstracted robot maps based on the robots' scan fields in (a). Each circle represents a robot map and the overlapping area means the two robots' scan fields have intersection (i.e., data redundancy between the robot maps). ]{
        \begin{minipage}[t]{\linewidth}
        \centering
        \includegraphics[width=0.65\linewidth]{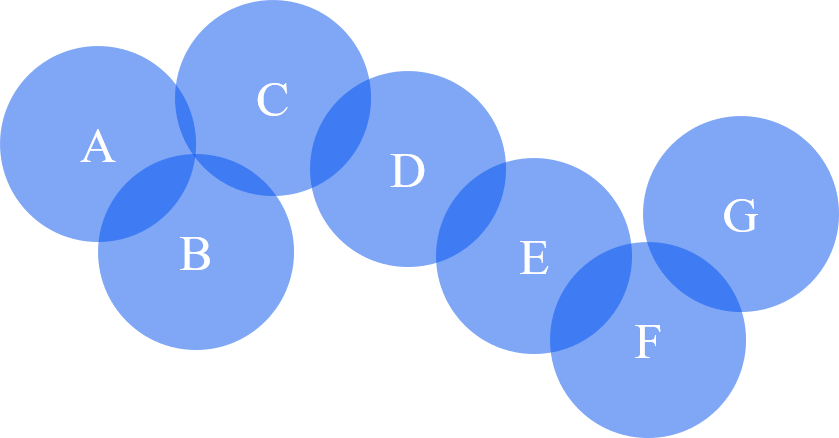}
        \end{minipage}
    }
    \subfigure[The corresponding graph of the maps in (b). A vertex is a robot map and a link is the overlapping relation between robot maps, with the overlapping degree as its weight.]{
        \begin{minipage}[t]{\linewidth}
        \centering
        \includegraphics[width=0.6\linewidth]{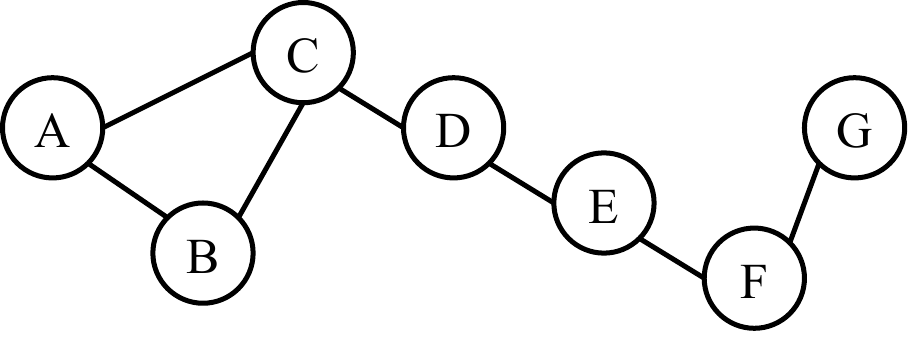}
        \end{minipage}
    }
    \caption{An example abstraction flow of the overlapping graph. For the robots in the factory scene (a), we first abstract robot maps in (b) and then derive the overlapping graph in (c).}
    \label{figure:overlapping_graph}
\end{figure}

\renewcommand{\algorithmicrequire}{\textbf{Input:}}  
\renewcommand{\algorithmicensure}{\textbf{Output:}} 

\begin{algorithm}[t]
    \caption{Balanced robot data grouping algorithm} 
    \label{algorithm:grouping}
    \begin{algorithmic}[1]
        \REQUIRE The overlapping graph $\mathcal{G}$ (with $V$ vertices), \\
        The number of edge servers $N$, \\
        The neighboring function $\mathcal{N}_C(v)$ that returns the neighbor vertices of $v$ in a set $C$
        \ENSURE Grouping result $\langle P_1, P_2, \cdots, P_N \rangle$ 
        \STATE Initialize an empty buffer set $R$ and an candidate set $C$
        \STATE Assign all vertices in $\mathcal{G}$ to $C$, set exit $flag = 0$ 
        \STATE Initialize empty grouping result $\langle P_1, P_2, \cdots, P_N \rangle$

        \REPEAT
            \IF {$R$ is empty}
                \FOR {$i = 1, 2, \cdots, N-1$} 
                    \STATE Select vertex $v^{\text{neigh}}_{\text{min}}$ from $C$ such that $v^{\text{neigh}}_{\text{min}}$ has the least number of neighbors
                    \STATE $P_i \leftarrow P_i + \{v^{\text{neigh}}_{\text{min}}\}$ 
                    \STATE $C \leftarrow C - \{v^{\text{neigh}}_{\text{min}}\}$\\
                    \STATE $R \leftarrow R + C \cap \mathcal{N}_C(v^{\text{neigh}}_{\text{min}}) - \{v^{\text{neigh}}_{\text{min}}\}$
                    \IF {$|C| \leq \frac{V}{N}$}
                        \STATE set $flag = 1$ and break
                    \ENDIF
                \ENDFOR
        \ELSE
                \FOR {$i = 1, 2, \cdots, N-1$}  
                    \STATE Select vertex $v^{\text{gain}}_{\text{max}}$ from $C$ such that $v^{\text{gain}}_{\text{max}}$ has the maximum gain using Equation (\ref{eq:gain})
                    \STATE $P_i \leftarrow P_i + \{v^{\text{gain}}_{\text{max}}\}$
                    \STATE $C \leftarrow C-\{v^{\text{gain}}_{\text{max}}\}$
                    \STATE $R \leftarrow R+ C \cap \mathcal{N}_C(v^{\text{gain}}_{\text{max}}) - \{v^{\text{gain}}_{\text{max}}\}$\\
                    \IF {$|C| \leq \frac{V}{N}$}
                        \STATE set $flag = 1$ and break
                    \ENDIF
                \ENDFOR
            \ENDIF
        \UNTIL {$flag = 1$}
        \STATE $P_N \leftarrow C$ 
        \STATE Apply Algorithm 2 to optimize the grouping result
        \RETURN $\langle P_1, P_2, \cdots, P_N \rangle$
    \end{algorithmic}
\end{algorithm}

\subsection{Balanced Robot Data Grouping}
The robot data grouping targets at dividing the $V$ robots' SLAM sensory data into $N$ groups (i.e., $N$ edge servers) with the objective of evenness.
Here evenness means the number of robots across groups should be approximately equal, in order to achieve balanced processing workloads among the edge servers later on. 
Besides, for minimizing the size of edge maps after fusion, the robot data within the same group is desired to yield robot maps with a higher overlapping degree as much as possible.
We accomplish these goals by firstly abstracting an overlapping graph to shape the relations between robot maps.

\textbf{Graph abstraction.}
Fig. \ref{figure:overlapping_graph} illustrates how we abstract an overlapping graph.
Given the robots in Fig. \ref{figure:overlapping_graph}(a)'s factory layout plane, we can collect their position and localization information in a bottom-up manner: the robots report their meta sensory data to the nearby edge servers and the edges push these data to the cloud to infer the global overlapping information.
In many well-planned deployments, this procedure can be even accomplished offline as long as we know the robots' scanning routes prior to runtime.
With the information, we can therefore represent the robots' corresponding maps in Fig. \ref{figure:overlapping_graph}(b), where each one reflects the area that a robot's scanning field covers.
It is represented by circle, because the robot can scan the surrounding 360-degree during the movement.
Let a pair of maps with an overlapping area share a link, we abstract a graph in Fig. \ref{figure:overlapping_graph}(c), named as the overlapping graph.
A vertex in such a graph is a robot map, whereas a link quantifies the overlapping degree between two robot maps. 
Particularly, two vertices are not connected if there is no overlap between the two robot maps.
We denote the overlapping degree as $w(u,v)$, which represents the weight of the link between vertex $u$ and $v$.

\textbf{Balanced grouping algorithm.}
Using the overlapping graph model, we observe that the robot map grouping problem can be approximated to a graph clustering variant.
Concretely, we can operate on the overlapping graph to decide a balanced grouping with maximal in-group overlapping degree\footnote{Particularly, in-group overlapping degree means the geometrical sum of $w(u,v)$ within the same group, while out-group degree is the sum of $w(u,v)$ between two specific groups.}. 
To accomplish this, we develop a balanced grouping solution in Algorithm \ref{algorithm:grouping} on the basis of the balanced graph partitioning method  \cite{wu2014algorithms,andreev2006balanced}.
The key idea is first to generate a feasible grouping plan and next use Tabu searching for further optimization.
Particularly, the way that Algorithm \ref{algorithm:grouping} finds an initial grouping is to iteratively put the vertex with maximum performance gain to a group such that for each group the in-group overlapping degree is maximized, and the Tabu searching optimization is to avoid a local optimum result and will be described in Algorithm \ref{algorithm:tabu} detailedly.

\begin{algorithm}[t]
  \caption{Tabu-searching-based optimization} 
  \label{algorithm:tabu}
  \begin{algorithmic}[1]
  \REQUIRE The overlapping graph $\mathcal{G}$, \\
  Initial grouping $\langle P_1, P_2, \cdots, P_N \rangle$,\\
  The maximum iterations times $\kappa$, \\
  The maximum tabu list length $\tau$
  \ENSURE Optimized grouping result $\langle P^*_1, P^*_2, \cdots, P^*_N \rangle$
  \STATE Initialize an empty tabu list $T$ of vertex pairs, an empty list $G$ of obtained grouping results, and an empty list $S$ of vertex pairs to be swapped
  \STATE $\langle P^*_1, P^*_2, \cdots, P^*_N \rangle \leftarrow \langle P_1, P_2, \cdots, P_N \rangle$
    \REPEAT
        \FOR{vertices $v, u \in \mathcal{G}$}
            \IF{$v$ and $u$ are not in the same group}
                \STATE Add $(v, u)$ to $S$
                \STATE Swap the locations of $v$ and $u$ to obtain a new grouping $\langle P'_1, P'_2, \cdots, P'_N \rangle$
                \STATE Add $\langle P'_1, P'_2, \cdots, P'_N \rangle$ to $G$
            \ENDIF
        \ENDFOR
        \STATE $k \leftarrow \arg\min\limits_{i,g_i \in G}\{f(g_i)\}$ using Equation (\ref{eq:fitness})
        \IF {$f(G_k) < f(\langle P^*_1, P^*_2, \cdots, P^*_N \rangle)$}
            \IF {$S_k \in T$}
                \STATE Remove $S_k$ from T
            \ENDIF
            \STATE $\langle P^*_1, P^*_2, \cdots, P^*_N \rangle \leftarrow G_k$
        \ENDIF
        \IF {$S_k \notin T$}
            \STATE Add $S_k$ to $T$
        \ENDIF
        \IF {$|T| > \tau$}
            \STATE Remove the first element of $T$ 
        \ENDIF
        \STATE $\langle P_1, P_2, \cdots, P_N \rangle \leftarrow G_k$
    \UNTIL {Iterating for $\kappa$ times}
    \RETURN $\langle P^*_1, P^*_2, \cdots, P^*_N \rangle$
  \end{algorithmic}
\end{algorithm}

The input of Algorithm \ref{algorithm:grouping} are an undirected graph $\mathcal{G}$ (with $V$ vertices), the number of groups (i.e., the number of edge servers) $N$, and a neighboring function $\mathcal{N}(v)$ that returns the neighbor vertices of $v$.
The output is the grouping result $\langle P_1, P_2, \cdots, P_N\rangle$, where $P_i$ is a set that includes the vertices assigned to $i$-th partition.

At the beginning of Algorithm \ref{algorithm:grouping}, we initialize an empty buffer set $R$ and an empty candidate set $C$, and immediately add all vertices in the overlapping graph $\mathcal{G}$ to $C$.
We also initialize empty grouping result $\langle P_1, P_2, \cdots, P_N\rangle$, and then goes into an iterating process.
For each iteration, we first check whether the buffer $R$ is empty: if empty, it means there is no suitable candidate related to vertices in the current group $P_i$, thus we add the vertex $v^{\text{neigh}}_{\text{min}}$ with the least number of neighbors from the candidate set $C$, and put it to $P_i$ (Line 7-9).
The vertices associating with $v^{\text{neigh}}_{\text{min}}$, obtained by $\mathcal{N}(v)$, will be added to $R$ for the coming assignment (Line 10).
Whenever a node is deleted from $C$, we check whether {$|C| \leq \frac{V}{N}$}. If the condition is satisfied, which means the difference of nodes' number in any two groups is less than 2, it will set the exit $flag$ as true and jump out of the loop (Line 11-13).
It should be noted that these neighboring vertices $\mathcal{N}(v)$ have never been visited, i.e., they have not been assigned to a specific group.
If $R$ is not empty, there remains vertices in the buffer, and we will select the vertex $v^{\text{gain}}_{\text{max}}$ with the maximum gain from candidates $C$ and move it to $P_i$. The gain is estimated using Equation (\ref{eq:gain}): given a vertex $v$ in $C$, we reckon the sum of $w(u,v)$ in $P_i$ and $C$, respectively, and calculate their difference.
Such difference values actually measures the weight change brought by the vertex migration.
The greater the difference, the greater the possibility of the point being selected.
\begin{align}
    \text{gain}(v)=\sum_{u \in P_{i}} w(u, v)-\sum_{u \in C\backslash\{v\}} w(u, v).
    \label{eq:gain}
\end{align}

The principle behind exploiting the gain formula is to make each in-group overlapping degree as large as possible, while the out-group is as small as possible.
Again, we assign $v^{\text{gain}}_{\text{max}}$ to the current group $P_i$ and remove it from $C$, and the neighboring vertices $\mathcal{N}(v^{\text{gain}}_{\text{max}})$ will be added to the buffer $R$ (Line 17-20).
In this step, we still need to check whether {$|C| \leq \frac{V}{N}$} is met (Line 21-23).
The above iterating processing continues until exit $flag$ is true, which indicates that all the vertices have been allocated in a balanced manner.
The remaining vertices in $C$ will be directly assigned to $N$-th group $P_N$, and therefore we obtain a feasible initial grouping result (Line 26).

At the same time, $\mathcal{N}(v)$ will be added to $R$. Finally, when all the vertices are visited, we can get an appropriate grouping result (Line 27).
However, this result is based on heuristic assignment and may terminate at a local optimum. 
To avoid such circumstances, we further apply a Tabu searching to seek a better grouping result (Line 28), as described in Algorithm \ref{algorithm:tabu}.

\textbf{Tabu-searching-based optimization.}
The key idea of Algorithm \ref{algorithm:tabu} is to use a tabu list to save the local optimal solution, to avoid falling into the same result.
The input of the algorithm is the overlapping graph $\mathcal{G}$, the initial grouping result $\langle P_1, P_2, \cdots, P_N \rangle$, the maximum iterations times $\kappa$ and the maximum tabu list length $\tau$.
The output is an optimized grouping result $\langle P^*_1, P^*_2, \cdots, P^*_N \rangle$.

Algorithm \ref{algorithm:tabu} begins with initializing an empty tabu list $T$ with length $\tau$, an empty list $G$ of obtained grouping results, and an empty list $S$ of vertex pairs to be swapped.
We instantiate the expected grouping $\langle P^*_1, P^*_2, \cdots, P^*_N \rangle$ with the initial grouping $\langle P_1, P_2, \cdots, P_N \rangle$, and goes into an iterating process.
For each iteration, we first construct all possible swapping pairs (Line 4-10): find vertices $v$ and $u$ that are not in the same group, add them to $S$, apply a swapping to obtain an updated grouping $\langle P'_1, P'_2, \cdots, P'_N \rangle$, and put the updated grouping to $G$.
With these swapped candidates, we then expect to filter those that improve the quality of grouping.
Specifically, we define a fitness function $f(\cdot)$ in Equation (\ref{eq:fitness}):
\begin{align}
    f(\langle P_1, P_2, \cdots, P_N \rangle)=\sum_{u \in P_i, v \in P_j, i \neq j} w(u, v).
    \label{eq:fitness}
\end{align}

The \textit{fitness} defines the suitability of the grouping by calculating the sum of weighted edges between different groups in Equation (\ref{eq:fitness}). 
It is designed for weighing the sum of edges in an overlapping graph, and we can therefore use it in our optimization on workload grouping.
The smaller the fitness value, the better the grouping result. 
In Algorithm \ref{algorithm:tabu}, we find the swapping vertex pair, noted with index $k$, such that it has the smallest fitness within all candidates in $G$ (Line 11).
If this vertex pair $S_k$ has smaller fitness than the current optimized counterpart's ($\langle P^*_1, P^*_2, \cdots, P^*_N \rangle$), we will update this optimized result (if it is exactly in the tabu list $T$, just remove it from $T$).
Otherwise, $S_k$ is accepted and will be added $S_k$ to the tabu list (Line 18-20), which means we will accept the local optimal solution and prevent it from being visited again, unless the amnesty rules are met.
If the tabu list $T$ is overlength than the preset maximum length $\tau$, we will remove the first element, the one that is the most out-of-date, from $T$, based on the first-in-first-out principle.
The current grouping will also be updated will $k$-th grouping.
The above iterations continue until reaching the maximum $\kappa$ times, and the final optimized grouping $\langle P^*_1, P^*_2, \cdots, P^*_N \rangle$ will be returned.

\textbf{Complexity analysis.}
Assuming an overlapping graph has $V$ vertices and $L$ links and we need to divide $N$ groups, the time complexity of the balanced grouping algorithm is $O(V+L)$. In the worst case, each partition has to traverse $V/N$ vertices. The operation of selecting vertices is regarded as $O(1)$, because this part can be optimized by maintaining the tree on the data structure. Moreover, the optimization requires traversing the edges, so the total complexity is $O(V/N+L)=O(V+L)$.

\begin{algorithm}[t]
    \caption{Resource-aware edge offloading algorithm} 
    \label{algorithm:offloading}
    \begin{algorithmic}[1]
        \REQUIRE The grouping result $\langle P_1, P_2, \cdots, P_N \rangle$,\\
        The edge servers list $\mathcal{E} = \langle 1, 2, \cdots, N\rangle$, \\
        Bandwidth $B_j$ between edge server $j$ and the cloud, \\
        Latency prediction model $L_j(P_i)$ that returns the computing latency of $P_i$ on edge server $j$,\\
        The function $\mathcal{D}(P_i)$ that returns the output data size of partition $P_i$
        \ENSURE  Offloading plan $\langle \theta_1, \theta_2, \cdots, \theta_N \rangle$
        \FOR{$i = 1, 2, \cdots, N$}
            \STATE $j \leftarrow \arg \min \limits_{j \in \mathcal{E}} (L_j(P_i) + \frac{\mathcal{D}(P_i)}{B_j})$
            \STATE $\theta_i \leftarrow j$
            \STATE Remove edge server $j$ from $\mathcal{E}$
        \ENDFOR
        \RETURN $\langle \theta_1, \theta_2, \cdots, \theta_N \rangle$
    \end{algorithmic}
\end{algorithm}

\subsection{Resource-Aware Edge Offloading}
After robot data grouping, we have several groups of robot data and need to decide their placement to the proper edge servers for edge map fusions to minimize the total collaborative processing time.
Since the data transmission from the robots and edge servers can be a bottleneck and the network conditions of the edge servers can be different, the decision needs to tailor for the available bandwidth.
Moreover, the computing capacity of the edge servers can also be heterogeneous. 
Therefore, we propose a resource-aware edge offloading strategy in Algorithm 3.

Algorithm 3 works in a greedy fashion and the input is the robot data grouping result $\langle P_1, P_2, \cdots, P_N \rangle$ from Algorithm \ref{algorithm:grouping}, the edge servers list $\mathcal{E} = \langle 1, 2, \cdots, N\rangle$, and $B_j$ that records the bandwidths between edge server $j$ and the cloud.
Besides, two functions are also employed: the latency prediction function $L_j(P_i)$ that returns the computing latency of $P_i$ on edge server $j$, the function $\mathcal{D}(P_i)$ that returns the data size of partition $P_i$.
The output is the offloading strategy $\langle \theta_1, \theta_2, \cdots, \theta_N \rangle$, where $\theta_i$ logs the edge server that grouping $i$ will be placed.
In Algorithm \ref{algorithm:offloading}, we iteratively select a group $P_i$ and pick an edge server $j$ such that the robot-to-edge offloading (including transmission and computation) latency is minimized (Line 2).
Particularly, we use $L_j(P_i)$ to estimate the computing latency on the edge server and $\frac{\mathcal{D}(P_i)}{B_j}$ to calculate the transmission time.
Since this latency simultaneously tackles computing aspect, i.e., $L_j(P_i)$, and the bandwidth $B_j$, it is aware of the available resource in both computation and communication, and can support heterogeneous cases.
The selected edge server $j$ that minimizes the latency will be assigned to $\theta_i$ and be removed from the available edge server list $\mathcal{E}$.
We carry out this process until all edge servers are assigned with a group, and return the final offloading plan $\langle \theta_1, \theta_2, \cdots, \theta_N \rangle$.

\textbf{Complexity analysis.}
In our implementation, we use an offline profiling method to obtain the estimation of computing time and output data size, which will be detailed in Section \ref{sec:implementation}.
Using the profiling result, the overhead of calculating the computing/transmission time of each server is $O(1)$, the complexity of Algorithm 3 is $O(N)$, which is linear and fast.

\section{Implementation}
\label{sec:implementation}

\begin{figure}[t]
    \centering
    \includegraphics[height=4.2cm]{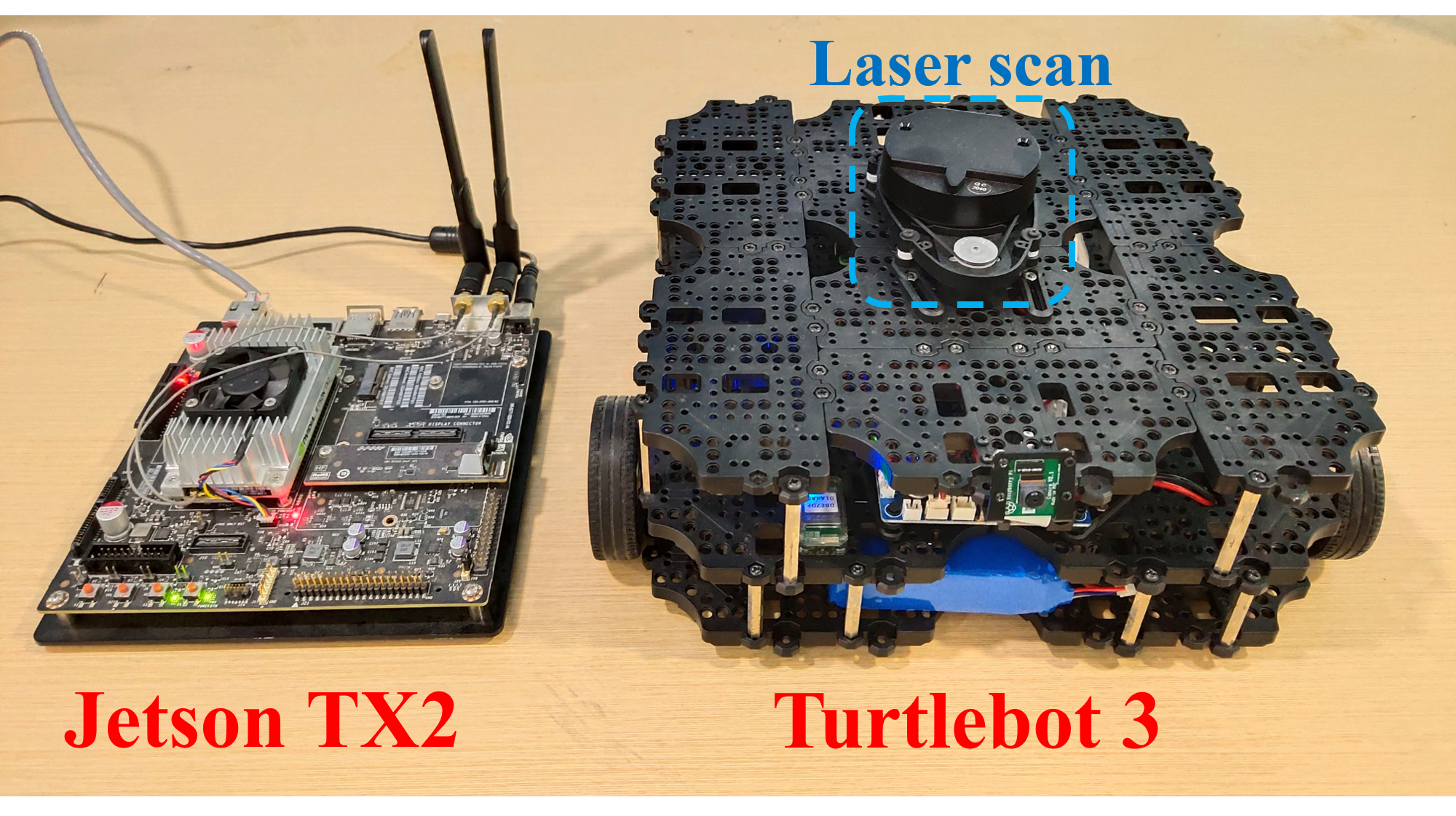}
    \caption{Our prototype employs Turtlebot 3 as the robot and Jetson TX2 as the edge server. The Turtlebot is equipped with a RPLIDAR A1 as the laser scan and a Raspberry Pi as the processing module. }
    \label{fig:hardware}
\end{figure}

\begin{figure}[t]
    \centering
    \includegraphics[height=4.2cm]{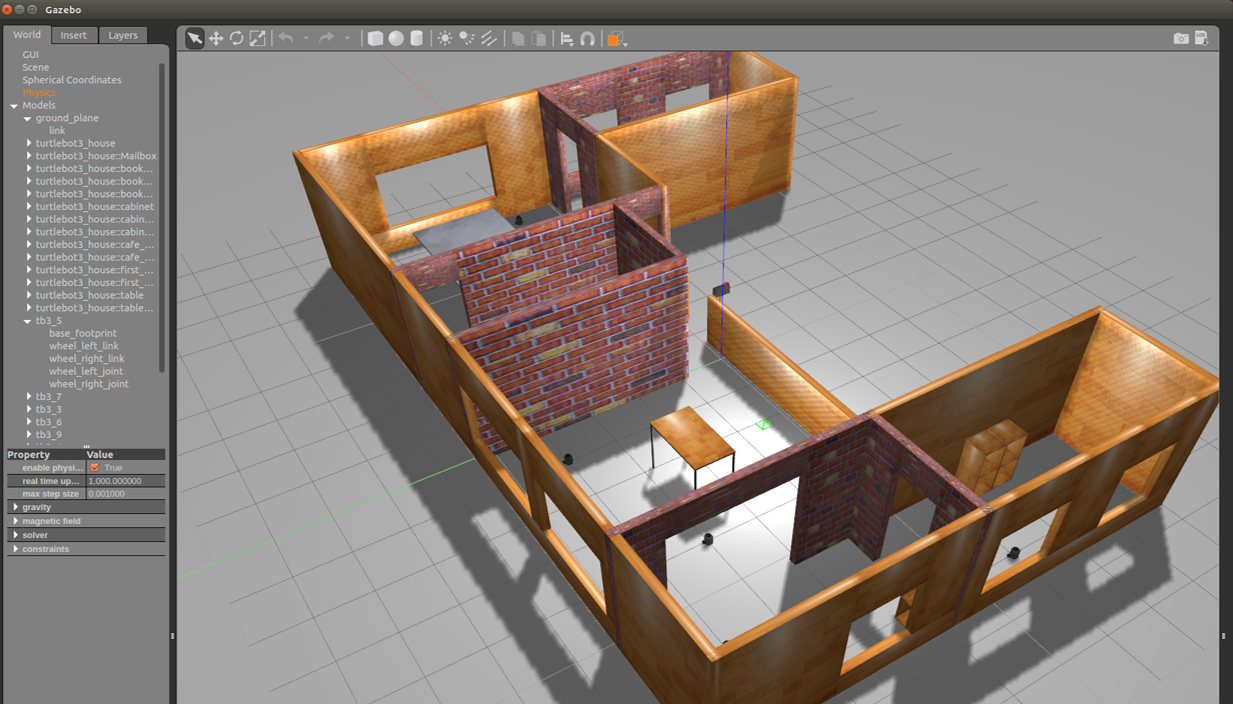}
    \caption{The simulated scene of Fig. \ref{figure:simulation}(a) in the Gazebo \cite{koenig2004design} simulator, where robots and edge servers are placed inside to emulate multi-robot SLAM applications in physical edge scenarios.}
    \label{fig:gazebo_gui}
\end{figure}

We have implemented RecSLAM in both realistic prototype and simulation platform.
The proof-of-concept prototype is built with Jetson TX2 and Turtlebot 3 as shown in Fig. \ref{fig:hardware}, while the simulations are carried on Gazebo\cite{koenig2004design}, a high-precision simulator of robotics emulations.
Fig. \ref{fig:gazebo_gui} shows the graphical interface of the scene in Fig. \ref{figure:simulation}(a).

\textbf{SLAM integration.}
We adopt GMapping as the SLAM kernel on edge servers. However, our system can support different SLAM toolkits as long as the robot can provide the necessary data. 
Each time a robot offloads data to the edge, the edge server will create a dedicated GMapping instance for the robot to calculate the local map with a specific topic.
For example, when two robots offload data to an edge server at the same time, the topics of them are \textsf{tb4\_0} and \textsf{tb4\_2} respectively. At this time, there will be two GMapping instances on the edge that subscribe the data with \textsf{tb4\_0/} and \textsf{tb4\_2/} namespace, and then publish the occupancy grid map with \textsf{tb4\_0/map} and \textsf{tb4\_2/map} topic respectively.

To leverage the proposed scheduling algorithms over GMapping, we implement them in a centralized manner. 
Specifically, we select one of the available edge servers as the coordinator to be responsible for collecting robots’ maps information, estimating the
overlapping degrees, and running scheduling algorithms. 
According to the scheduling result, the coordinator will route the robot data to corresponding edge servers, and orchestrate them to collaboratively run map fusion tasks.
The algorithms can also be implemented distributionally, where a tailored mechanism is desired for global information sharing and consensus management, and we leave it as a future work.

\textbf{Robot-edge communication.}
The communications among robots, edge servers, and the coordinator are conducted via ROS\footnote{ROS is an operating system specifically designed for manipulating robotics. It provides a loosely coupled distributed communication framework and rich libraries of various functional packages, making it convenient to carry out robotic system development.}. They all communicate using the publisher-subscriber model. In such a mechanism, the subscriber node will first register on the node master, and then the node master will search among the registered publishers, find the matching publisher, and finally the publisher and the subscriber will establish a communication channel.

The communications involve multiple subscribing topics. In RecSLAM, the data format released by the robot cluster is \textsf{tbn\_x}, where \textsf{x} is the specific robot number. The data format of the GMapping instance subscripter in edge is \textsf{tbn\_x}, and the data with topic \textsf{tbn\_x/map} will be published after processing. The Map Fusion module receives \textsf{coon\_x/map} and publishes the data with \textsf{edgem\_x/map} at the same time.

For coordinator, it will receive \textsf{tbn\_x/map}'s data. The workload grouping algorithm in the coordinator will split the robots into groups, and then the offloading algorithm will generate allocation decisions and publish the data with the topic \textsf{coon\_x/map}. The communication mechanism based on ROS ensures the reliability of communication between robot and edge.

\textbf{Parameter profiling.}
The coordinator derives the overlapping degree matrix using a modified package called \textsf{weight\_cal}. It is based on \textsf{multirobot\_map\_merge}, a generic package in ROS.
It can read multiple maps in a batch, extract their features, perform matching checking, and outputs a rough overlapping degree matrix.
To run it in a lightweight style, we invoke this package periodically, i.e., calculates and saves the overlapping degree matrix of the robot maps in a preset regular frequency.

\begin{figure}[t]
    \centering
    \includegraphics[width=0.8\linewidth]{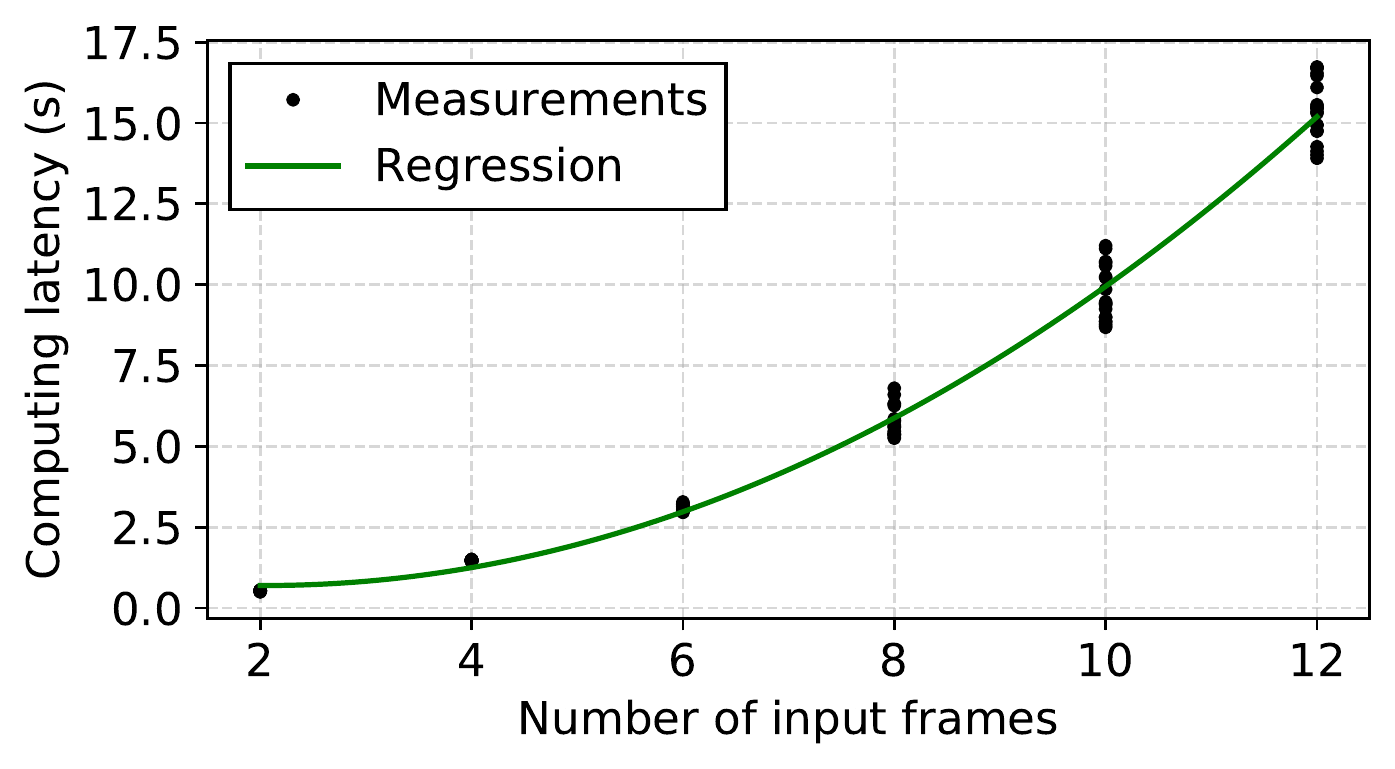}
    \caption{The regression result of SLAM computing latency on the Jetson TX2. The dots are the measurements and the curve is the regression.}
    \label{fig:regression_model}
\end{figure}

To estimate the time of preparative fusion $t^{\text{fuse}}_e$ at the edge server, we employ an offline profiling mechanism that trains regression models for performance prediction.
In our deployment, we carry out profiling on Turtlebot 3 and record the computing latency of map fusion with respect to the number of data frames.
Fig. \ref{fig:regression_model} illustrates the regression result.
An interesting observation of the figure is that the computation latency rises in a quadratic trend as the frame number increases from 2 to 12, 
This is because the computation of merging requires pairwise matching of data frames, and the number of matches has an exponential relationship with the number of frames.

For the transmission latency between robots, edge servers and the cloud, there exists many network bandwidth measure tools available, which enables the real-time transmission time estimation based on the data size of the input frames. 
In RecSLAM, we utilize the principle of tolerance and exclusion formula to estimate the output size of preparative fusion.
Since the number of robot maps is relatively small, the estimation error of the overlapping area is within a tolerable range.

\section{Evaluation}
\label{sec:evaluation}

\begin{figure*}[t]
    \centering
    \subfigure[Prototype scene.]{
        \begin{minipage}[t]{0.2\textwidth}
        \centering
        \includegraphics[height=3.0cm]{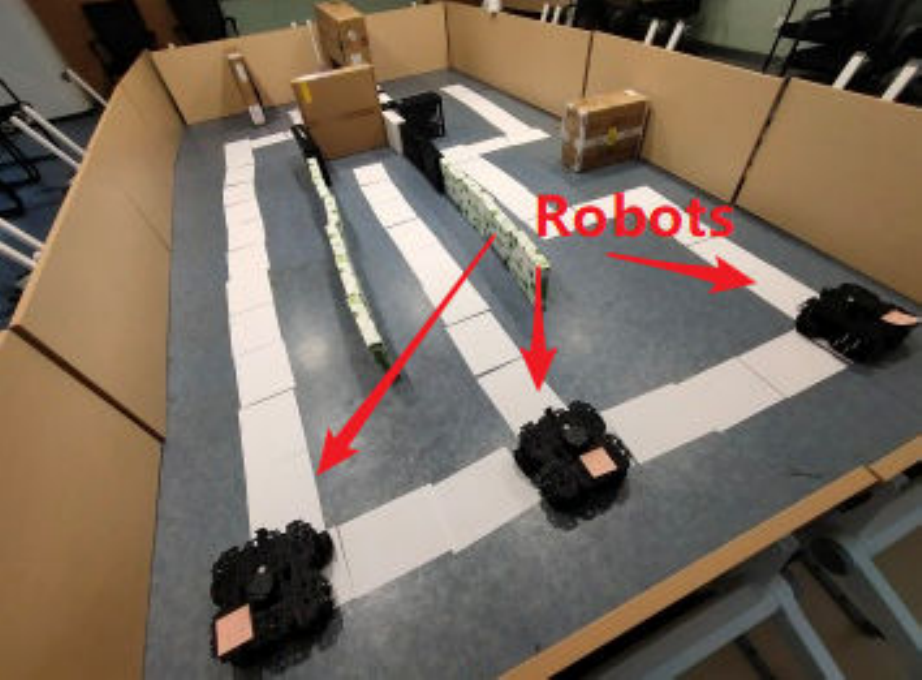}
        \end{minipage}
    }
    \quad
    \subfigure[Robot map 1.]{
        \begin{minipage}[t]{0.12\textwidth}
        \centering
        \includegraphics[height=3.0cm]{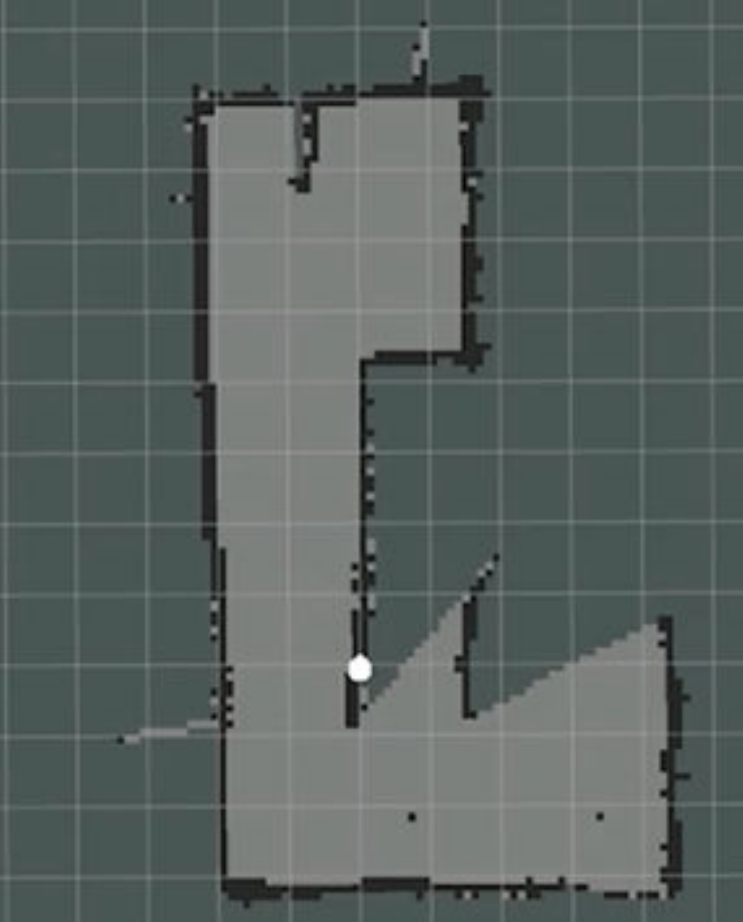}
        \end{minipage}
    }
    \quad
    \subfigure[Robot map 2.]{
        \begin{minipage}[t]{0.12\textwidth}
        \centering
        \includegraphics[height=3.0cm]{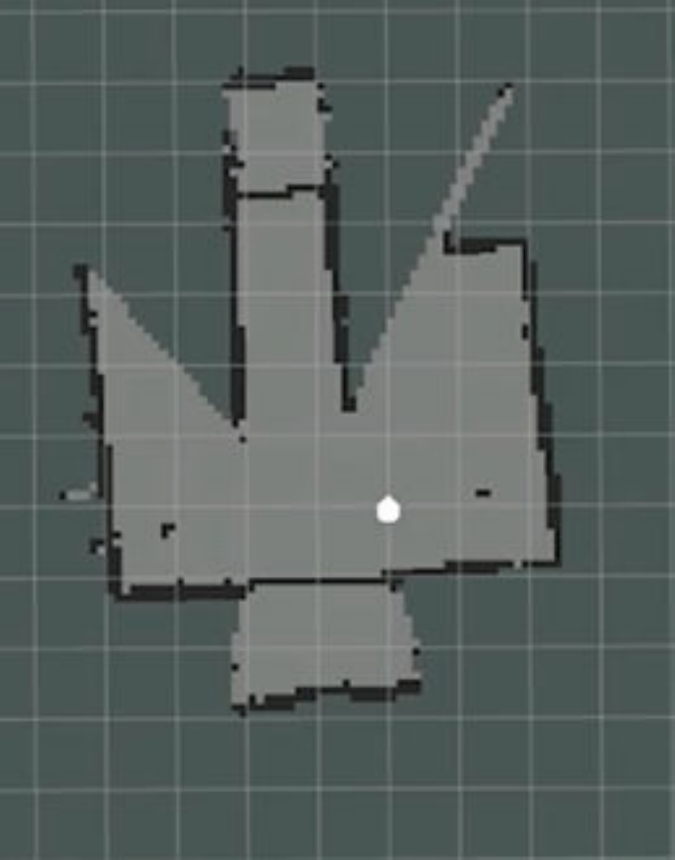}
        \end{minipage}
      }
    \quad
    \subfigure[Robot map 3.]{
        \begin{minipage}[t]{0.12\textwidth}
        \centering
        \includegraphics[height=3.0cm]{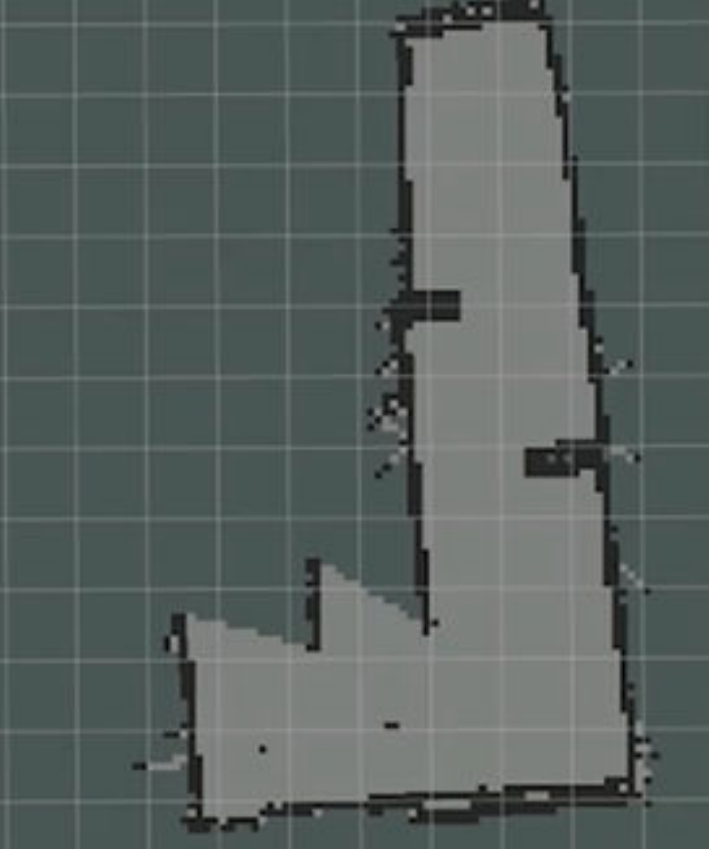}
        \end{minipage}
    }
    \quad
    \subfigure[Edge map.]{
        \begin{minipage}[t]{0.2\textwidth}
        \centering
        \includegraphics[height=3.0cm]{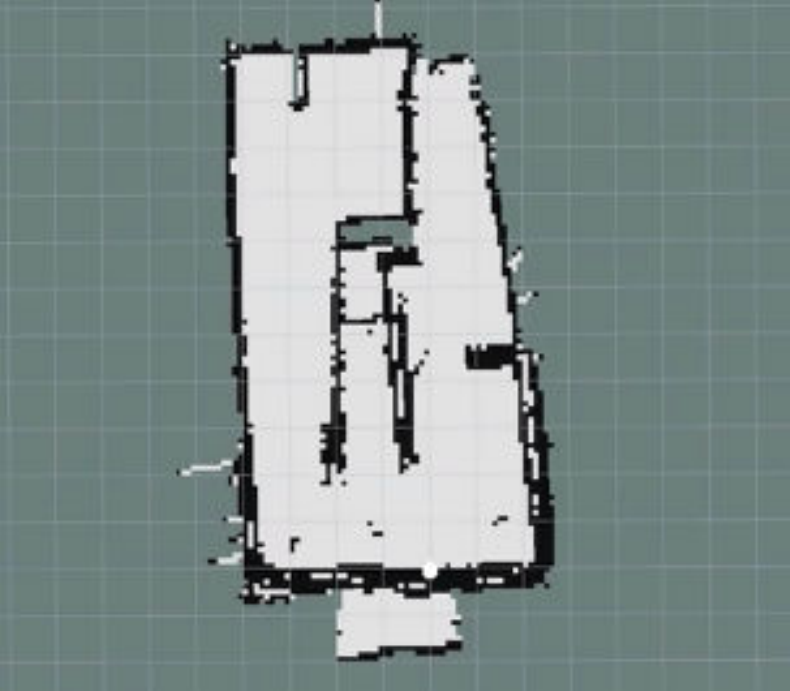}
        \end{minipage}
    }
    \caption{Proof-of-concept prototype experiments with three robots and one edge server in a self-built real scene. Due to the limitations in robotics hardware and physical space, the testbed in (a) is not built perfectly and gaps exist between walls, for which the corresponding global map in (e) has a rough boundary. Despite that, the global map still shows its feasibility by sketching the complete scene and reversing the obstacles' details, which verifies the practicality of RecSLAM.}
    \label{figure:real_deployment}
\end{figure*}

\begin{figure*}[t]
    \centering
    \subfigure[Scene.]{
        \begin{minipage}[t]{0.1\textwidth}
        \centering
        \includegraphics[height=1.8cm]{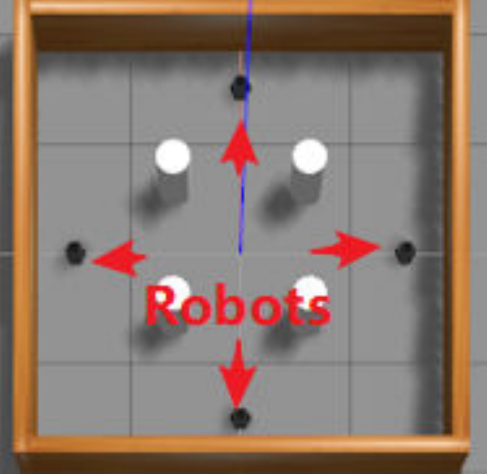}
        \end{minipage}
    }
    \subfigure[Robot map 1.]{
        \begin{minipage}[t]{0.1\textwidth}
        \centering
        \includegraphics[height=1.8cm]{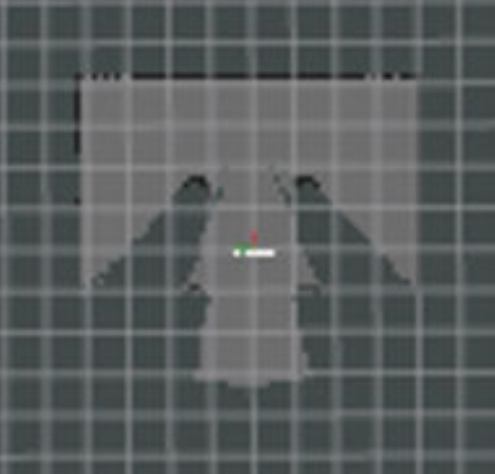}
        \end{minipage}
    }
    \subfigure[Robot map 2.]{
        \begin{minipage}[t]{0.1\textwidth}
        \centering
        \includegraphics[height=1.8cm]{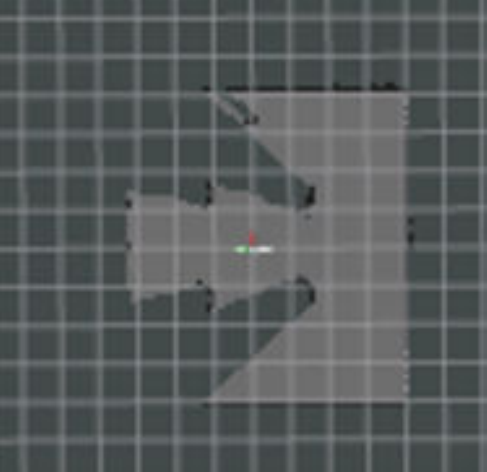}
        \end{minipage}
    }
    \subfigure[Robot map 3.]{
        \begin{minipage}[t]{0.1\textwidth}
        \centering
        \includegraphics[height=1.8cm]{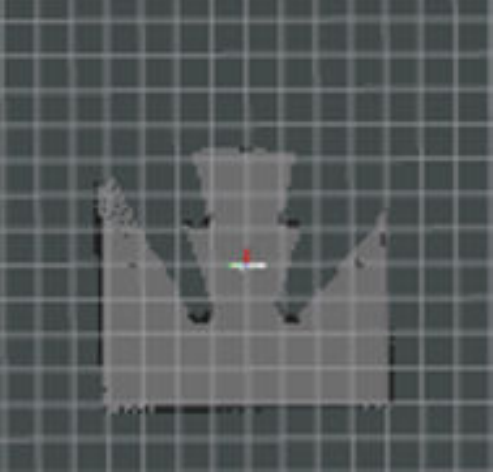}
        \end{minipage}
    }
    \subfigure[Robot map 4.]{
        \begin{minipage}[t]{0.1\textwidth}
        \centering
        \includegraphics[height=1.8cm]{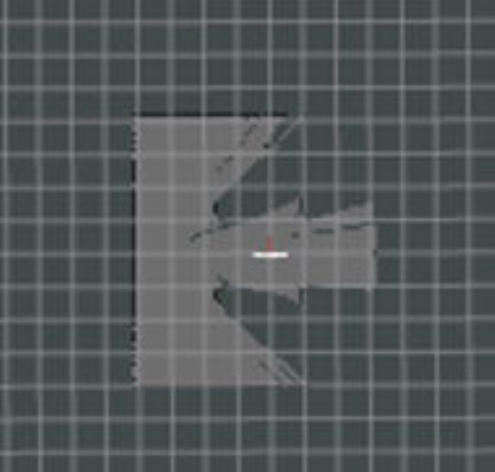}
        \end{minipage}
    }
    \subfigure[Edge map 1.]{
        \begin{minipage}[t]{0.1\textwidth}
        \centering
        \includegraphics[height=1.8cm]{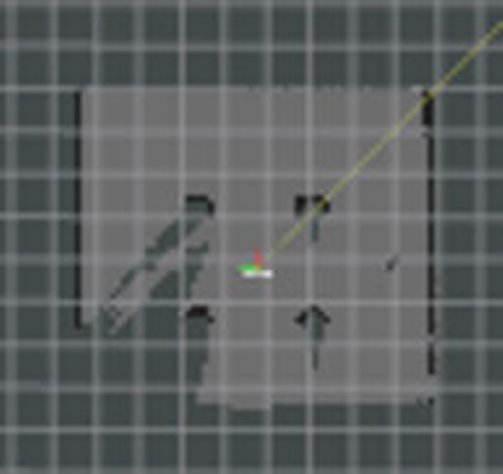}
        \end{minipage}
    }
    \subfigure[Edge map 2.]{
        \begin{minipage}[t]{0.1\textwidth}
        \centering
        \includegraphics[height=1.8cm]{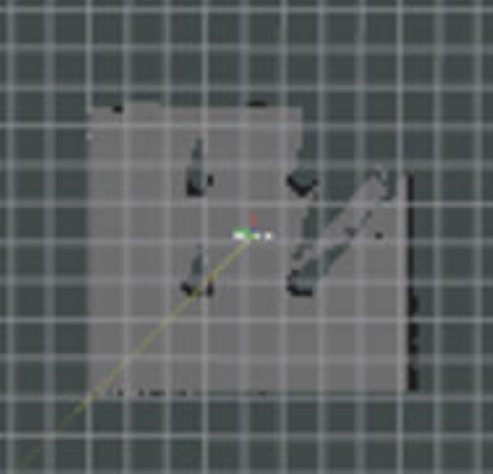}
        \end{minipage}
    }
    \subfigure[Global map.]{
        \begin{minipage}[t]{0.1\textwidth}
        \centering
        \includegraphics[height=1.8cm]{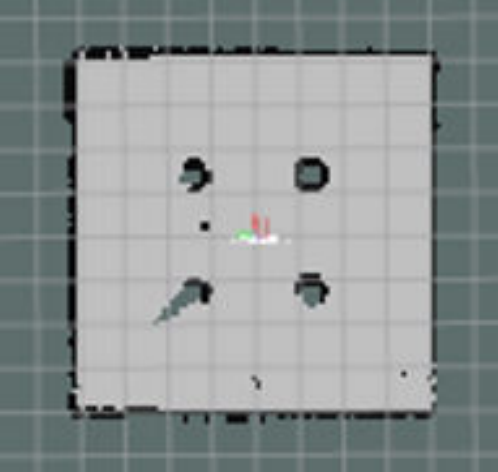}
        \end{minipage}
    }
    \caption{Simulation results with four robots and two edge servers in a symmetrical apartment scene. The edge servers are not drawn in (a) for simplicity. The robots walk according to preset routes and pass their sensory data to edge servers for multi-robot SLAM computation. The complete global map demonstrates RecSLAM's effectiveness in merging multi-robot sources.}
    \label{figure:simulation_scene2}
\end{figure*}

\begin{figure*}[t]
    \centering
    \subfigure[Scene.]{
        \begin{minipage}[t]{0.1\textwidth}
        \centering
        \includegraphics[height=1.8cm]{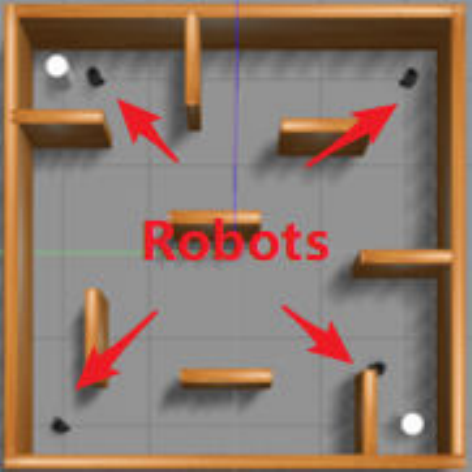}
        \end{minipage}
    }
    \subfigure[Robot map 1.]{
        \begin{minipage}[t]{0.1\textwidth}
        \centering
        \includegraphics[height=1.8cm]{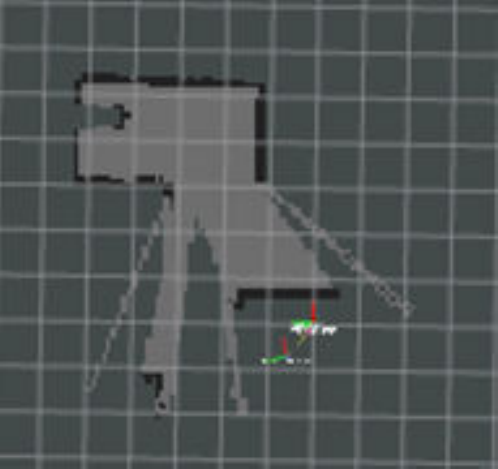}
        \end{minipage}
    }
    \subfigure[Robot map 2.]{
        \begin{minipage}[t]{0.1\textwidth}
        \centering
        \includegraphics[height=1.8cm]{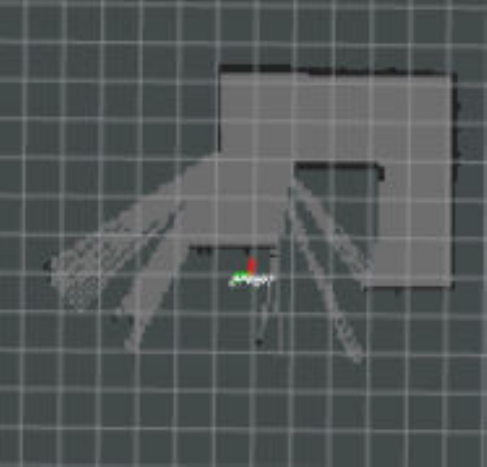}
        \end{minipage}
    }
    \subfigure[Robot map 3.]{
        \begin{minipage}[t]{0.1\textwidth}
        \centering
        \includegraphics[height=1.8cm]{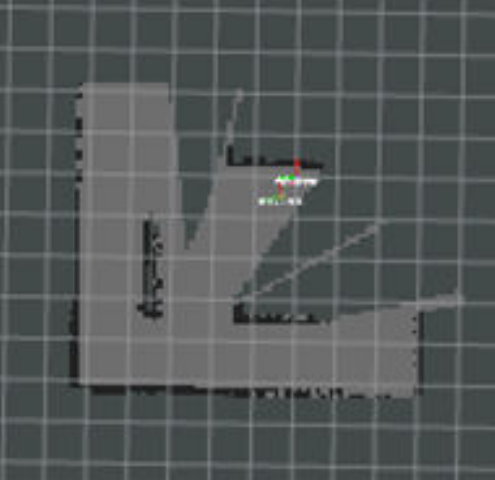}
        \end{minipage}
    }
    \subfigure[Robot map 4.]{
        \begin{minipage}[t]{0.1\textwidth}
        \centering
        \includegraphics[height=1.8cm]{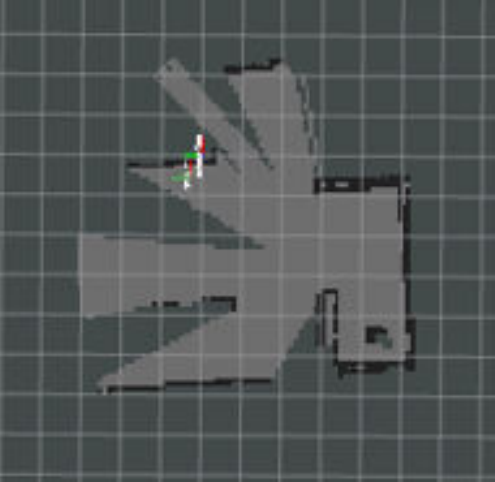}
        \end{minipage}
    }
    \subfigure[Edge map 1.]{
        \begin{minipage}[t]{0.1\textwidth}
        \centering
        \includegraphics[height=1.8cm]{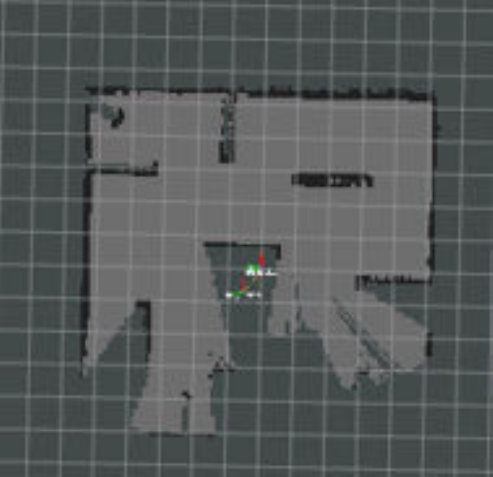}
        \end{minipage}
    }
    \subfigure[Edge map 2.]{
        \begin{minipage}[t]{0.1\textwidth}
        \centering
        \includegraphics[height=1.8cm]{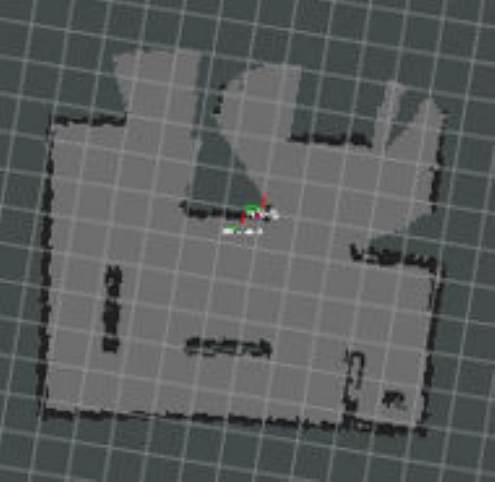}
        \end{minipage}
    }
    \subfigure[Global map.]{
        \begin{minipage}[t]{0.1\textwidth}
        \centering
        \includegraphics[height=1.8cm]{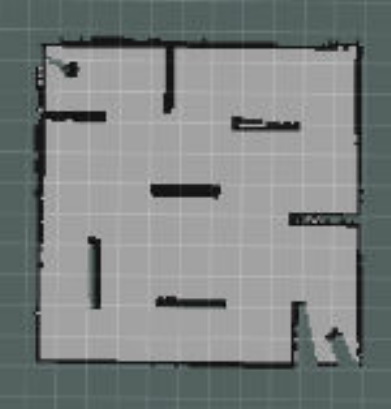}
        \end{minipage}
    }
    \caption{Simulation results with four robots and two edge servers in an asymmetrical apartment scene. The edge servers are not drawn in (a) for simplicity. RecSLAM effectively renders the complete mapping again, and note that the flaws in the global map mean the area that the robots have not scanned.}
    \label{figure:simulation_scene3}
\end{figure*}

This section evaluates RecSLAM in both a simulator platform and a prototype deployment.
Particularly, we focus on the feasibility of global map construction and the end-to-end execution time of running a multi-robot SLAM process.

\begin{figure*}[t]
    \centering
    \begin{minipage}[t]{0.3\textwidth}
    \centering
        \includegraphics[height=5cm]{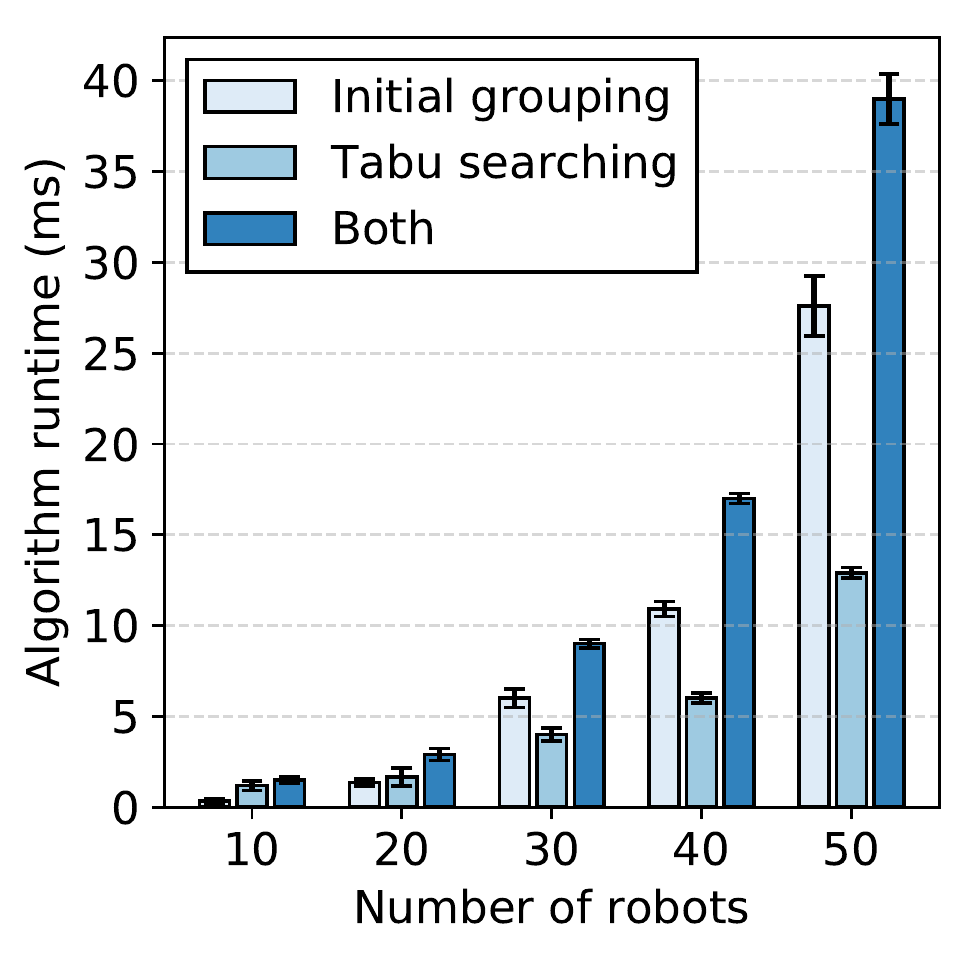}
        \caption{The algorithm runtime of different components in RecSLAM with varying number of simulated robots.}
        \label{fig:evaluation_ablated}
    \end{minipage}
    \qquad
    \begin{minipage}[t]{0.3\textwidth}
        \includegraphics[height=5cm]{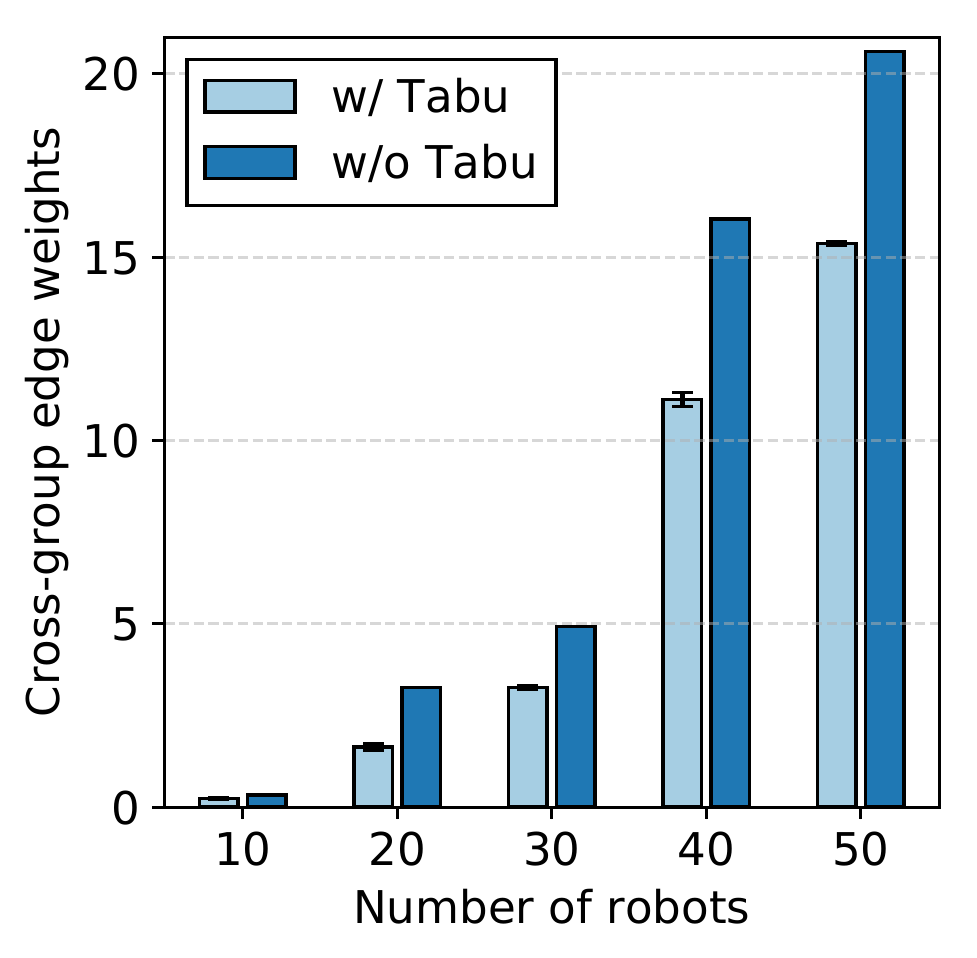}
        \caption{The optimization effect of RecSLAM's grouping algorithm with and without Tabu optimization. Lower cross-group edge weights is better.}
        \label{fig:tabu_cut}
    \end{minipage}
    \qquad
    \begin{minipage}[t]{0.3\textwidth}
    \centering
        \includegraphics[height=5cm]{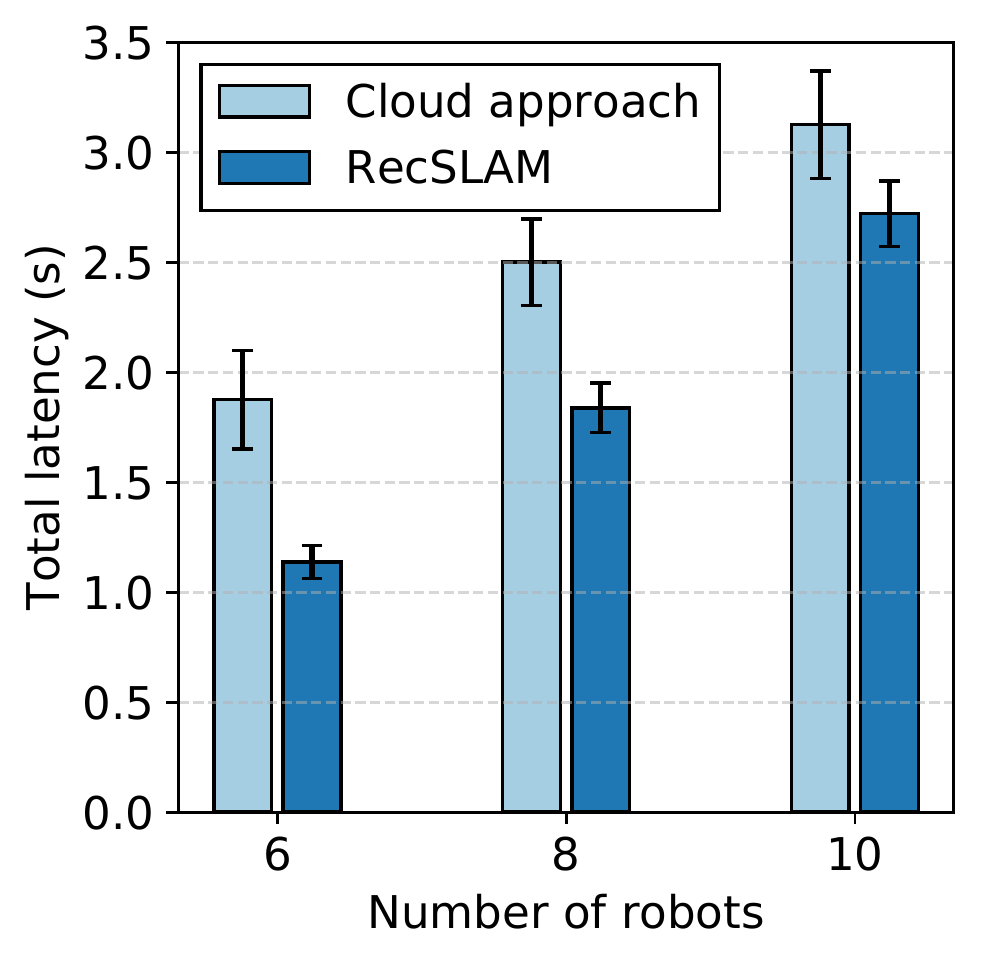}
        \caption{Total execution time of RecSLAM and the state-of-the-art cloud approach under WiFi environment.}
        \label{fig:evaluation_total_latency}
    \end{minipage}
\end{figure*}

\begin{figure*}[t]
    \centering
    \begin{minipage}[t]{0.3\textwidth}
        \centering
        \includegraphics[height=5cm]{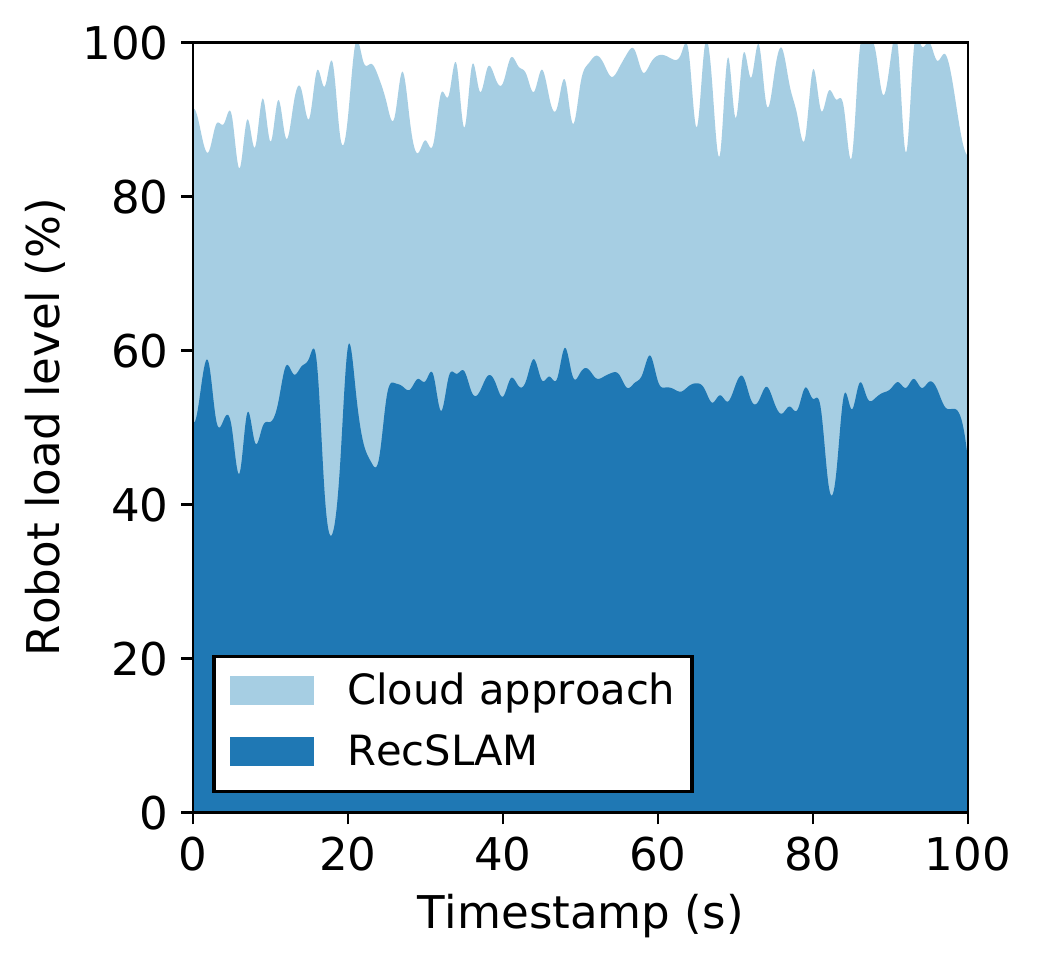}
        \caption{The recorded load level of the robot under the cloud approach and RecSLAM.}
        \label{fig:evaluate_cpu_workload}
    \end{minipage}
    \qquad
    \begin{minipage}[t]{0.3\textwidth}
        \centering
         \includegraphics[height=5cm]{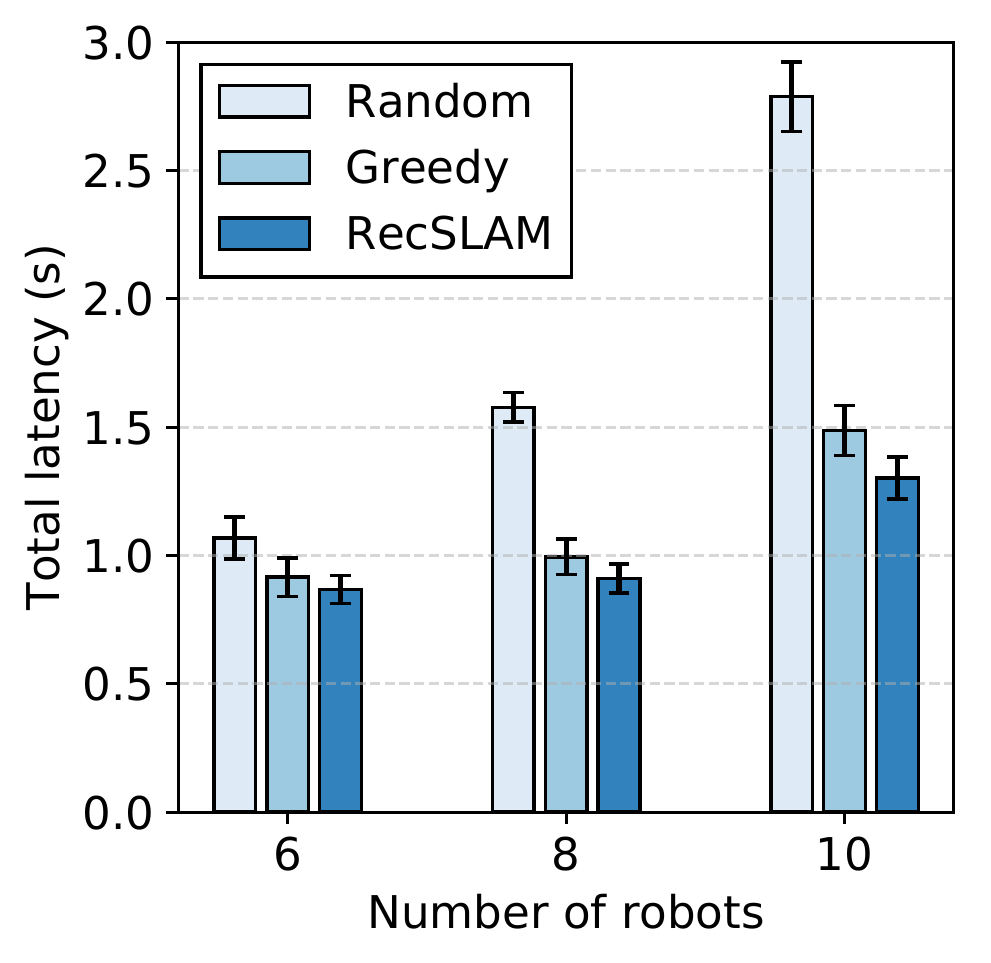}
        \caption{Total execution time of RecSLAM and ablated scheduling policies with varying number of robots.}
        \label{fig:evaluation_scheduling_compare}
    \end{minipage}
    \qquad
    \begin{minipage}[t]{0.3\textwidth}
        \centering
        \includegraphics[height=5cm]{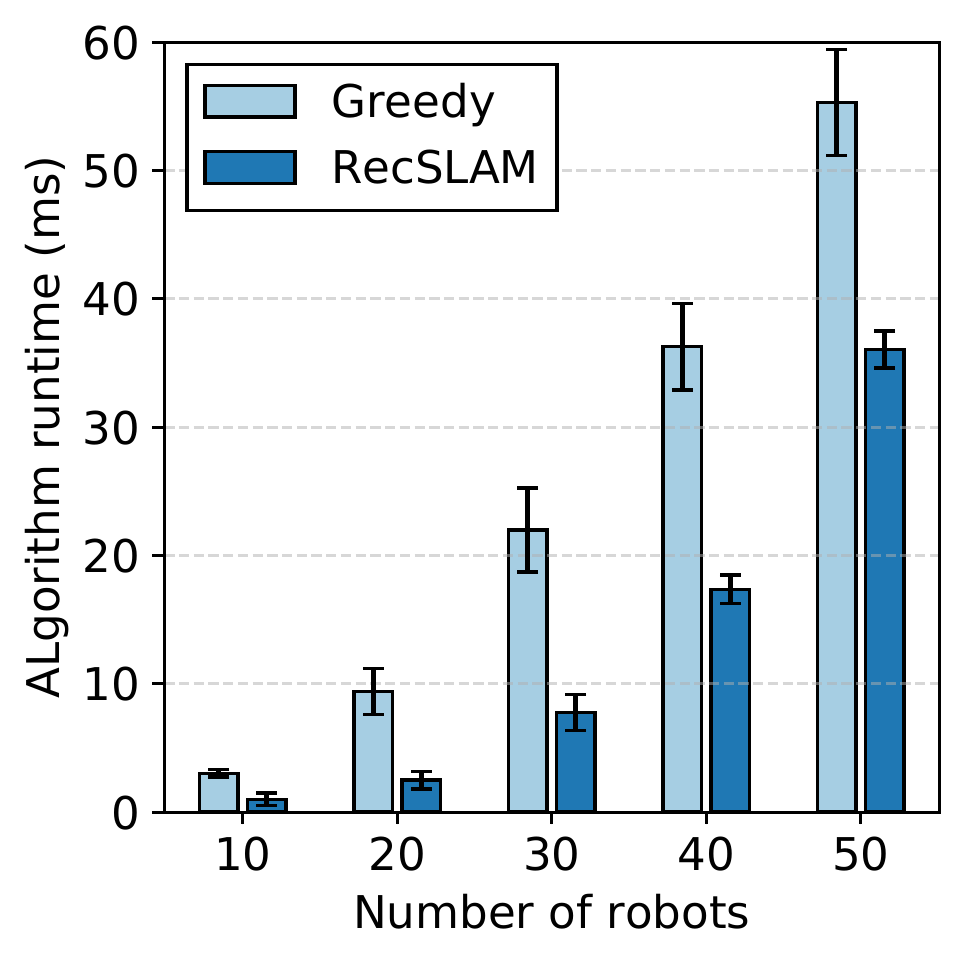}
        \caption{The algorithm runtime of RecSLAM and the greedy counterpart with varying number of robots.}
        \label{fig:evaluation_scheduling_overhead}
    \end{minipage}
\end{figure*}

\subsection{Functional Verification}

We first conduct functional verification of RecSLAM based on a small-scale realistic prototype.
Specifically, we mainly focus on collecting realistic robotics measurements and examining the effectiveness of the hierarchical map fusion procedure, and thus providing necessary data to facilitate subsequent larger-scale simulation experiments.
We use the TurtleBot3 Waffle Pi mobile robot for laser data collection and the Jetson TX2 as the edge server. 
The TurtleBots are all installed with Raspberry Pi 3b+ as their processing module, which has 1.4GHz ARM Cortex-A53 CPUs and 1GB LPDDR2 memory.
The Jetson TX2 is equipped with NVIDIA Denver and ARM Cortex-A57 CPUs, an NVIDIA Pascal GPU with 256 CUDA-cores, and 8 GB LPDDR4 memory.
All the robots and the edge server are connected via the 802.11ac wireless signal and in the same LAN. 
The cloud server employs an instance (8vCPU, 16GB memory, Ubuntu 16.04) that locates at the same available zone with the robots and their physical distance is about 200km.
The testbed scene is built in a $6\times5.5m^2$ conference room, where we manually add walls and obstacles, as shown in Fig. \ref{figure:real_deployment}.
The linear velocity of the robots is $0.20m/s$, and the angular velocity is set to $0.20-0.80m/s^2$ during the runtime. 
Although RecSLAM's design allows robots to walk freely, we have set determined routes for robots in advance for the sake of simplicity and experimental repeatability.
As the robots move around, the equipped laser sensors perceive the surrounding environment at a fixed frequency and continue collecting sensory data.

Fig. \ref{figure:real_deployment} presents the mapping results of our prototype.
Particularly, Fig. \ref{figure:real_deployment}(b), (c) and (d) visualize the robot maps that are computed based on the robots' sensory data, where we can see that the robots pave three different routes on the left, center and right of the scene, respectively.
Overlapping areas can be observed from these robot maps, e.g., robot map 1 in (b) and robot map 2 in (c) share the same area at their bottom.
Fig. \ref{figure:real_deployment}(e) shows the obtained edge map that is fused from the robot maps, where the complete scene is sketched and the obstacles' details are reserved.

To further demonstrate RecSLAM's effectiveness, we make additional simulation experiments and build several virtual scenes in Gazebo \cite{koenig2004design}, a popular platform\footnote{Gazebo \cite{koenig2004design} is a robotics simulator that offers the ability to accurately and efficiently simulate populations of robots in complex indoor and outdoor environments, and has been widely used in the community.} that provides high-precision and near-physical experimental environments.
Specifically, we design typical apartment indoor scenes in Gazebo's virtual environments and place robots and edge servers to emulate multi-robot scenarios at the network edge.
In all scenes, we distribute and preset routes for robots in advance, and drive them to walk around the apartment.
During the runtime of multi-robot SLAM, we record the robots' sensory data and calculate the maps by continuously replaying the collected data.
To fully reflect the specifications in our prototype, we configure the same motion parameters, such as velocity and scanning field, and replay the robots' movement for all simulations.

The first scene, as shown in Fig. \ref{figure:simulation}(a), is the main scene for our simulation experiments, which is a $13\times18m^2$ apartment scene with three edge servers (and a varying number of robots on demand).
Fig. \ref{figure:simulation}(b) is exactly the global map generated by RecSLAM.
We also built another two scenes in Fig. \ref{figure:simulation_scene2} and Fig. \ref{figure:simulation_scene3} to examine the effectiveness in constructing the whole map.
The scene in Fig. \ref{figure:simulation_scene2}(a) is in a $4\times4m^2$ square and is designed to emulate a symmetrical layout.
The subfigures Fig. \ref{figure:simulation_scene2}(b)-(e) depicts the robot maps constructed using the laser scan data from the four robots, while the subfigures (f) and (g) are the corresponding edge maps and (h) is the global map in the cloud.
Specifically, (f) is obtained by fusing (b) and (c), and (g) is from (d) and (e).
The last scene is in a $5\times5m^2$ room as shown in Fig. \ref{figure:simulation_scene3}(a), which exhibits an asymmetrical layout.
In contrast to the neat robot maps in Fig. \ref{figure:simulation_scene2}, the maps in Fig. \ref{figure:simulation_scene3} shows relatively irregular styles due to the scene's asymmetry.
Nevertheless, it can be seen from both scenes that the global map after map fusion matches the corresponding scenes with respect to the whole layout and the obstacle positions.

Remark that the global mapping results from RecSLAM and the cloud approach are the same since both of them reserve complete raw scan data from robots and run the same SLAM module. 
Particularly, RecSLAM improves multi-robot SLAM beyond traditional cloud approach by mitigating the workload of SLAM and fusion to edge servers and achieves much lower execution latency without sacrificing map accuracy.

\subsection{Performance Evaluation and Comparison}

Due to the limitations of the robotics hardware and physical space in our laboratory, we focus on verifying the functional feasibility of RecSLAM’s multi-robot map fusion.
To fully evaluate multi-robot SLAM performance with more robots and more edge servers, as well as in larger scenes, we conduct large-scale simulation experiments in Gazebo to demonstrate RecSLAM's effectiveness and efficiency.

\textbf{Algorithm runtime.}
RecSLAM consists of two parts, namely the grouping algorithm in Algorithm \ref{algorithm:grouping} and the Tabu-searching-based optimization algorithm in Algorithm \ref{algorithm:tabu}. 
In order to explore the algorithm's applicability in different situations, we inspect the performance of RecSLAM by breaking down the latency of grouping and Tabu searching in Fig. \ref{fig:evaluation_ablated}.
It should be noted that the undirected graphs generated in the test scenario before are all sparse matrices. 
Here we use randomly generated matrices for testing.

See Fig. \ref{fig:evaluation_ablated}, as the robot number increases, the total scheduling overhead climbs. In our scenario, tabu search always contributes the largest cost in the entire algorithm operation since the number of robots is small (< 20). But in dense scenarios, when the number of robots increases, we can find that the grouping overhead becomes larger, even exceeding tabu search. Because grouping is also a heuristic algorithm, it is greatly affected by the number of machines, but the algorithm runtime is still very small.

Since Tabu search overhead accounts for a relatively large amount of the algorithm runtime in our scenario, we further explored the impact of Tabu search though it is small. It can be seen from Fig. \ref{fig:tabu_cut} that with the number of robots from 10 to 50, Tabu search can obviously find a better solution, making the weights sum of the cross-group edges smaller.

\textbf{Performance comparison.}
Our first experiment targets at comparing RecSLAM with the cloud serving approach.
Fig. \ref{fig:evaluation_total_latency} shows their latency results with varying robot numbers, where RecSLAM enjoys lower latency across all settings.
Concretely, RecSLAM reduces 39.31\%, 26.50\% and 12.96\% latency with 6, 8 and 10 robots, respectively.
The result demonstrates the advantages came from edge servers' assists.
Specifically, the cloud approach lets robots compute SLAM locally and collects all robot maps in the cloud for centralized aggregation, where the workload on robots is always at a high level (Fig. \ref{fig:motivation_cpu_workload}).
In contrast, RecSLAM forces to transfer the SLAM workload and separate part of the fusion tasks to edge servers, therefore alleviating the computing stress on both sides.
As we measure during the RecSLAM runtime, the load level in robots records a mean of 53.91\% and a minimum of 36.18\%, significantly reducing the robot workload as shown in Fig. \ref{fig:evaluate_cpu_workload}.

To assess the effectiveness of RecSLAM's optimizing algorithms, we compare it with two heuristic counterparts and visualize the achieved latency results in Fig. \ref{fig:evaluation_scheduling_compare}, where RecSLAM refers to the corresponding scheduling policy mentioned above.
The random algorithm randomly distributes the robots' data to edge servers directly, recording the highest execution time among the three approaches.
Particularly, when serving 10 robots, it takes 2.79s, which is 2.15$\times$ higher than RecSLAM.
The greedy algorithm assigns the robot-edge data flow by first randomly initializing the grouping and next greedily selecting and swapping the robots in different groups.
For example, given $r$ robots and $n$ edge servers ($r\geq n$), it first selects any $r$ robots to respectively distribute to the $n$ groups.
For the rest operations, we orderly select and then swap robots in different groups such that the overlapping degree with the edge server's existing robot maps is maximized.
Such a mechanism partially utilizes the overlapping information, but still lacks a global determination on overall overlapping degrees.
As a result, it yields better partitioning than the random algorithm but is slower than RecSLAM.
RecSLAM considers the relations between robots' data by constructing an overlapping graph and scheduling the data flow in a global view.
Fig. \ref{fig:evaluation_scheduling_compare} shows that it reduces up to 53.41\% and 12.16\% latency over the random and greedy counterparts.

We further examine the algorithm runtime of RecSLAM and the greedy algorithm in Fig. \ref{fig:evaluation_scheduling_overhead} by simulating multiple robots in Gazebo\cite{koenig2004design} with numbers from 10 to 50.
In the figure, we observe that the RecSLAM consumes a smaller overhead than Greedy with an average of 58.71\% and up to 12.16\% lower latency.
Particularly, it takes less than 1ms for 10 robots and only 36ms for 50 robots, indicating that RecSLAM runs lightweight even with a large number of robots.

\begin{table*}[]
\centering
\caption{Comparing RecSLAM with relative edge-assisted SLAM frameworks.}
\label{table:related_work}
\begin{tabular}{lccccc}
\hline
\multicolumn{1}{c}{} & \textbf{Multi-robot support}    & \textbf{Multi-edge deployment}  & \textbf{Task decomposition}     & \textbf{Bandwidth adaptability}     & \textbf{Laser SLAM}             \\ \hline
CCM-SLAM \cite{schmuck2019ccm}                           & \cellcolor[HTML]{68CBD0}\Checkmark &                                 &                                 &                                 &                                 \\
EdgeSLAM \cite{xu2020edge}                           &                                 &                                 & \cellcolor[HTML]{68CBD0}\Checkmark & \cellcolor[HTML]{68CBD0}\Checkmark &                                 \\
Edge-SLAM \cite{ben2020edge}                          &                                 &                                 & \cellcolor[HTML]{68CBD0}\Checkmark &                                 &                                 \\
RecSLAM (Ours)                      & \cellcolor[HTML]{68CBD0}\Checkmark & \cellcolor[HTML]{68CBD0}\Checkmark & \cellcolor[HTML]{68CBD0}\Checkmark & \cellcolor[HTML]{68CBD0}\Checkmark & \cellcolor[HTML]{68CBD0}\Checkmark \\ \hline
\end{tabular}
\end{table*}

\section{Related Work}
\label{sec:related_work}

To the best of our knowledge, RecSLAM is the first one that optimizes multi-robot laser SLAM with the assist of edge computing. 
Only a few works are close to RecSLAM, and we list the most relative ones in Table \ref{table:related_work}.

\textbf{Multi-robot SLAM processing.}
The research on SLAM has experienced decades \cite{smith1986representation, smith1990estimating}. 
With years of development and practice, a number of single-robot SLAM algorithms \cite{davison2007monoslam, klein2007parallel, qin2018vins, labbe2019rtab, mur2017orb, mur2015orb, hess2016real, grisettiyz2005improving} has been proposed and deployed in real world.
For example, ORBSLAM2\cite{mur2017orb} can utilize a variety of sensors to collect image data and estimate pose during movement to achieve positioning and mapping, and has been used in underwater scenarios \cite{vargas2021robust,hidalgo2018monocular} and autonomous driving \cite{nobis2020persistent}.
Cartographer\cite{hess2016real} is a graph-optimized laser SLAM framework, which has been deployed on Google's street view services.

While these single-robot SLAM systems have been widely used, multi-robot SLAM remains to be intractable and still under exploration.
A considerable part of the works focuses on the algorithm layer, including multi-sensor fusion\cite{du2020multi,shao2019stereo,nguyen2021viral,palieri2020locus,zuo2020lic} and deep learning assistance\cite{kang2019df,wang2017deepvo,almalioglu2019ganvo}, etc. 
CCM-SLAM\cite{schmuck2019ccm} proposes a data-sharing mechanism between multiple robots to solve communication problems. CoSLAM\cite{zou2012coslam} proposes a collaborative SLAM system with multiple moving cameras in a possibly dynamic environment. CorbSLAM\cite{li2017corb} also uses a centralized method to detect the overlapping area of multiple local maps through DBoW method and uses global optimization through bundling adjustment to fuse these maps. 
However, a majority of these works implicitly rely on centralized servers, ignoring the practical scenarios of edge deployment.

\textbf{Edge-assisted SLAM computation.}
The emerging edge computing \cite{zhou2019edge,shi2016edge,mao2017survey} is a distributed computing paradigm that sinks the workload from the remote cloud to the servers in physically approximate to the end devices, and therefore significantly shortens the transmission distance and reduces the communication overhead.
In recent years, the community has been explored using edge computing to assist SLAM computation.
Dey et al.\cite{dey2016robotic} apply a particle filtering algorithm to an edge-cloud architecture and propose a dynamic offloading strategy.
However, it only accounts for a single robot scenario without considering transmission bottlenecks and realistic prototype implementation.
EdgeSLAM \cite{xu2020edge} propose an edge-assisted mobile semantic visual SLAM framework, which can perform visual SLAM in real-time. However, incorporating semantic segmentation brings great limitations to the system's scalability. Besides, it does not take multiple edge servers into consideration. 
Edge-SLAM \cite{ben2020edge} decouples SLAM from a modularity perspective, with the main focus on resource usage on mobile devices.
However, it only involves a single robot and an edge node, and cannot be directly extended to the communication between multi-robot and multi-edge.
While these systems intend to optimize SLAM computation from edge computing perspectives, they all focus on the single-robot scenario with peer-to-peer robot-edge synergy, ignoring the emerging multi-robot scenarios.
In contrast, our proposed system tackles a more complex SLAM deployment with the mutual presence of both multiple robots and multiple edge servers.
Besides developing a collaborative execution framework between robots and edge servers, we further design a coordinating policy to enable high-performance multi-robot SLAM map fusion.

\section{Conclusion}
\label{sec:conclusion}

This paper proposes RecSLAM, a multi-robot laser SLAM system under hierarchical robot-edge-cloud architecture. 
RecSLAM introduces edge servers to embrace the workload of SLAM computation and map fusion, which remarkably reduces the robot computing stress and saves the communication overhead between edge and cloud.
To optimize the overall performance, an efficient collaboration framework is further developed to orchestrate the data flow between robots and edge servers tailored to the heterogeneous network conditions.
Extensive evaluations with both simulation and real prototype implementation demonstrate the effectiveness and efficiency of RecSLAM over existing solutions.

\ifCLASSOPTIONcaptionsoff
  \newpage
\fi

\input{main.bbl}

\bibliographystyle{IEEEtran}
\bibliography{main}

\end{document}